%% file: main.tex
\documentclass{article}

\usepackage{microtype}
\usepackage{graphicx}
\usepackage{subcaption}
\usepackage{booktabs} 
\usepackage{hyperref}

\usepackage[accepted]{icml2026}
\usepackage{amsmath}
\usepackage{amssymb}
\usepackage{mathtools}
\usepackage{amsthm}
\usepackage{bbm}
\usepackage{bm}
\usepackage{multirow}
\usepackage{tocloft}
\usepackage{etoc}
\usepackage{appendix}
\usepackage[table]{xcolor}

\newcommand{\appentry}[2]{%
  \par\noindent\hyperref[#1]{#2}\dotfill\pageref{#1}\par
}

\usepackage[capitalize,noabbrev]{cleveref}
\theoremstyle{plain}

\theoremstyle{definition}

\theoremstyle{remark}

\usepackage[textsize=tiny]{todonotes}

\icmltitlerunning{Variational Adapter for Cross-modal Similarity Representation}

\begin{document}
\twocolumn[
  \icmltitle{Variational Adapter for Cross-modal Similarity Representation}




    \begin{icmlauthorlist}
      \icmlauthor{WenZhang Wei}{whu}
      \icmlauthor{Zhipeng Gui}{whu}
      \icmlauthor{Dehua Peng}{whu}
      \icmlauthor{Tiandi Ye}{ecnu}
      \icmlauthor{Huayi Wu}{lriesm}
    \end{icmlauthorlist}
    
    \icmlaffiliation{whu}{
    School of Remote Sensing and Information Engineering,
    Wuhan University, Wuhan 430079, China
    }
    
    \icmlaffiliation{ecnu}{
     School of Data Science and Engineering, East China Normal University, Shanghai 200062, China
    }
    
    \icmlaffiliation{lriesm}{
    State Key Laboratory of Information Engineering in Surveying, Mapping and Remote Sensing, Wuhan University, Wuhan 430079, China
    }

    \icmlcorrespondingauthor{Zhipeng Gui}{zhipeng.gui@whu.edu.cn}

  \icmlkeywords{Machine Learning, ICML}

  \vskip 0.3in
]



\printAffiliationsAndNotice{}  

\input{section/0_abstract}

\input{section/1_intro}
\input{section/2_related_work}
\input{section/3_methods}

\input{section/4_experiment}
\input{section/5_conclusion}


\section*{Acknowledgements}
This work is supported by the National Natural Science Foundation of China (No. 42571496, 42501573), Postdoctoral Fellowship Program and China Postdoctoral Science Foundation (No. BX20250084), China Postdoctoral Science Foundation (No. 2025M770345). We also thank Yuhan Huang for his helpful assistance with supplementary experiments and manuscript revision.

\section*{Impact Statement}
By enhancing the robustness of similarity measurement under imperfect supervision, this work contributes to the development of more trustworthy and cost-effective AI applications, particularly in scenarios with sparse annotations. However, modeling pairwise cross-modal similarity representations may introduce additional computational overhead, which should be considered in large-scale deployment.

\nocite{langley00}

\bibliography{paper}
\bibliographystyle{icml2026}

\newpage
\appendix
\onecolumn
\section*{\LARGE Appendix Contents}
\begingroup
\setlength{\parindent}{0pt}
\setlength{\parskip}{8pt plus 2pt minus 2pt}

\appentry{DBVP}{A. Related Work}
\appentry{STE}{B. Method Details}
\appentry{TaD}{C. Experimental Details}
\appentry{app:efficiency}{D. Computational Efficiency Analysis}
\appentry{sana}{E. Sensitivity Analysis}
\appentry{TLF}{F. Tolerance of Loss Functions}
\appentry{BNG}{G. Base-to-Novel Generalization}
\appentry{RUA}{H. Ranking-level Uncertainty Analysis}
\appentry{SD}{I. Similarity Distribution}
\appentry{VARR}{J. Visualization Analysis of Retrieval Results}
\endgroup
\input{section/6_appendix}

\end{document}

%% file: section/0_abstract.tex
\begin{abstract}
The core of vision-language models lies in measuring cross-modal similarity within a unified representation space. However, most image-text matching or multi-class image classification datasets lack fine-grained cross-modal matching annotations, forcing the continuous similarity space into binary classification boundaries. This compression induces false negative samples and significantly impairs the generalization performance of cross-modal tasks.
While prior research has attempted to mitigate this by modeling intra-modal ambiguity, it often overlooks inherent annotation flaws, leading to suboptimal uncertainty allocation. To address these challenges, we propose a Variational Adapter for Cross-modal Similarity Representation (VACSR). This approach reformulates image-text matching with fine-grained semantic scarcity as a variational inference problem. It constructs a latent space for cross-modal similarity and uses regularization techniques to mitigate overfitting to binary annotations.
Experiments on image-text retrieval, domain generalization, and base-to-novel generalization demonstrate the proposed method’s effectiveness and robust generalization ability.
\end{abstract}

%% file: section/1_intro.tex
\section{Introduction}
\label{sec:intro}
\begin{figure}[t]
\includegraphics[width=1.0\linewidth]{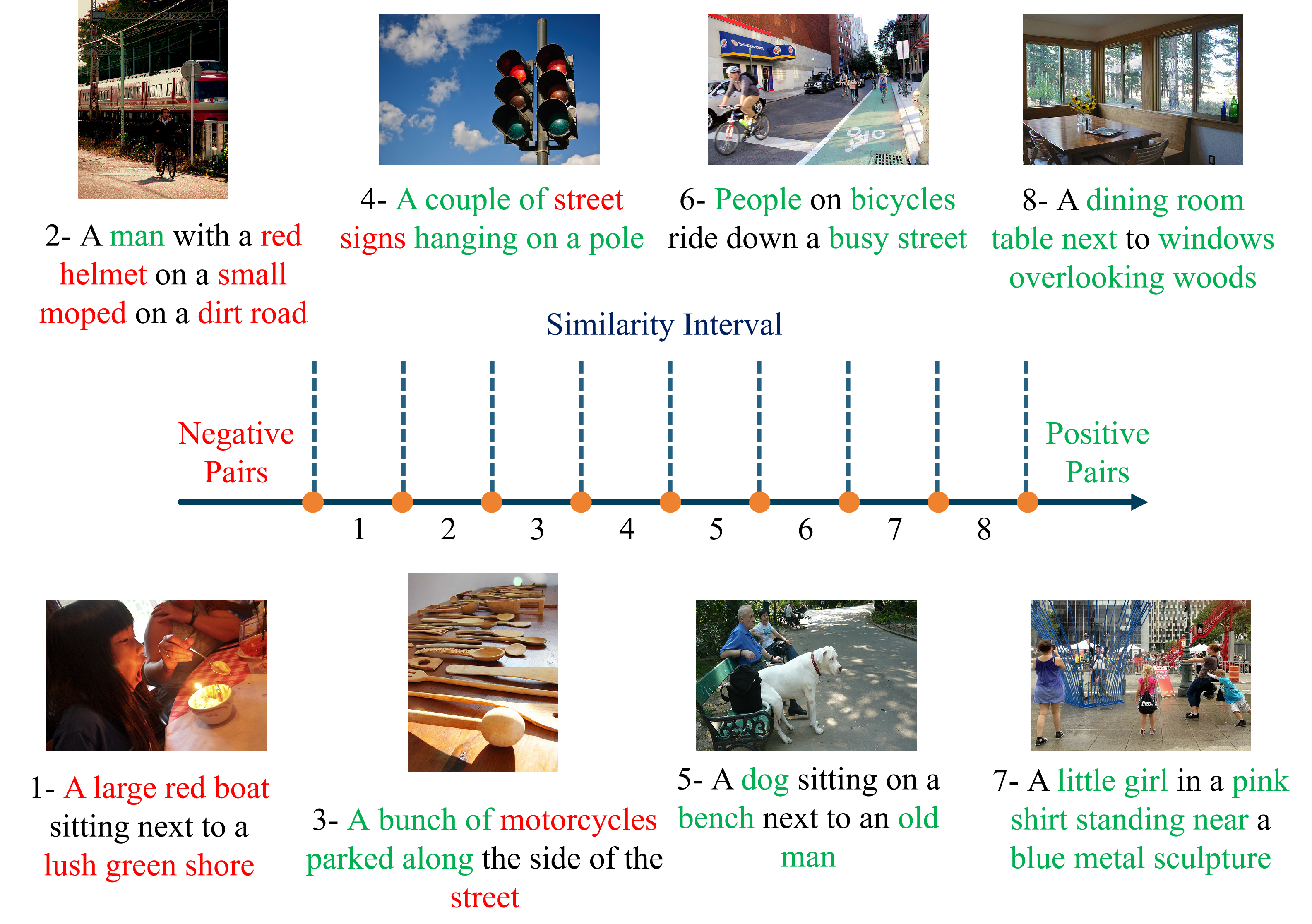}
\caption{Image-text pairs with varying levels of similarity. We computed the similarity of 40,000 sample pairs from the COCO Caption dataset and divided them into eight intervals in ascending order. From each interval, we randomly selected one image-text pair for visual presentation.}
\label{fig: pre}
\end{figure}
Unlike traditional visual recognition methods that depend on discrete labels and predefined visual concepts, vision-language models such as Contrastive Language-Image Pre-training (CLIP) \cite{radford2021learning, zhai2023sigmoid} directly align images and raw text within a shared representation space, thereby achieving impressive semantic understanding capabilities. These models have been applied to various downstream tasks, including zero-shot classification, cross-modal retrieval, and open-vocabulary detection \cite{pourpanah2022review, zhou2022learning, faghri2017vse++, chen2021learning, lee2018stacked, wei2025dynamic, wu2024towards}. However, multi-modal paired datasets like MS-COCO \cite{chen2015microsoft}, typically use a binary sparse annotation that labels image-text pairs as either ``matched" or ``mismatched." This labeling approach forces the model to strictly differentiate between all unannotated image-text pairs, potentially overlooking semantic relationships between samples. This issue is particularly evident in fine-tuning scenarios with limited samples, which can significantly compromise the model's generalization performance \cite{chun2022eccv, chun2024improved, gao2024clip}.

The matching relationship between image-text pairs (typically measured by similarity) is inherently complex. Consider an image of the Mona Lisa paired with the text ``The mysterious smile of the Mona Lisa": from an object co-occurrence perspective, they can be considered matched (the painting contains a smile), but the term ``mysterious" represents a subjective perception, making it challenging for models to assign an accurate similarity along this dimension. Furthermore, this relationship usually varies with similarity in a continuous and gradual pattern. As shown in Figure~\ref{fig: pre}, during the transition from samples annotated as negative to positive, the correspondence between images and texts becomes increasingly evident. For instance, although the third image does not fully match the text ``motorcycles", the objects within it are similarly ``parked along" the ground in ``a bunch". However, the binary sparse annotation lacks the necessary granularity to accurately measure the similarity of image-text pairs. This limitation results in the emergence of false negative sample pairs—i.e., pairs with a certain degree of semantic similarity that are incorrectly regarded as mismatched \cite{li2023integrating}. Previous studies have shown that FNs disrupt the semantic consistency of the representation space and limit the model's ability to capture complex matching relationships \cite{chun2021probabilistic}.

To mitigate the issues, existing work has proposed probabilistic embedding methods to model intra-modal ambiguity \cite{chun2022eccv, chun2024improved, li2023prototype, li2022differentiable, wang2022point, upadhyay2023probvlm, ji2023map, wei2025dynamic}. By mapping image and text data into random variables (instead of deterministic vector), these methods expand the potential set of matching results and construct a semantically rich retrieval space. However, such approaches still rely on binary sparse annotations when computing the cross-modal loss, implying that they address the false-negative pairs solely from the perspective of the data itself while overlooking annotation errors. As a result, the model would explain mislabeled false-negative pairs by inflating the uncertainty of individual samples, even when their semantic information is unambiguous. Although such uncertainty modeling can improve retrieval diversity, assigning excessive uncertainty to these samples could degrade retrieval precision.

To address the aforementioned problems, we propose a \textbf{Variational Adapter for Cross-modal Similarity Representation (VACSR)}, which models cross-modal similarity in a latent probabilistic space. Instead of directly fitting sparse binary annotations as deterministic targets, VACSR regularizes the similarity representation to recover continuous and smooth matching relationships. In this framework, uncertainty characterizes ambiguity in image-text matching rather than semantic ambiguity in individual samples. As a result, false-negative pairs (FNs) can be assigned higher uncertainty to reduce the effect of erroneous gradients induced by binary annotations, while positive pairs and informative hard-negative pairs are assigned lower uncertainty to strengthen the model’s discriminative capability. \textbf{A more detailed comparison between VACSR and probabilistic embedding methods is provided in Appendix~\ref{DBVP}.}

Experimental results demonstrate that VACSR achieves significant performance gains across various tasks, including image-text retrieval, noisy correspondence, domain generalization, and base-to-novel generalization. These findings indicate that VACSR exhibits strong robustness to noise while effectively improving model generalization and practical applicability in real-world settings.

%% file: section/2_related_work.tex
\section{Preliminary}
\label{PRELIMINARY}
For an image-text paired dataset $D=(X,Y)$, the conventional cross-modal alignment strategy employs separate feature encoders $\Phi(\cdot,\theta_{\mathcal{V}})$ and $\Psi(\cdot,\theta_{\mathcal{T}})$ to map paired data $(X_{i},Y_{j})$ into a d-dimensional space, obtaining their vector representations $\bm{v}_{i}=\Phi(X_{i},\theta_{\mathcal{V}}),\bm{t}_{j}=\Psi(Y_{j},\theta_{\mathcal{T}})$, where $\bm{v}_{i}, \bm{t}_{j}\in \mathbb{R}^{d}$. A pairwise loss is then typically applied to maximize the similarity of positive pairs while minimizing that of negative pairs, and the cross-modal similarity is measured using cosine similarity. This process is regarded as supervised learning and employs binary annotations to optimize the loss. (Hereafter, we make no distinction between the terms ``sample" and ``sample pair".)

\begin{figure*}[t]
\centering
\includegraphics[width=1\linewidth]{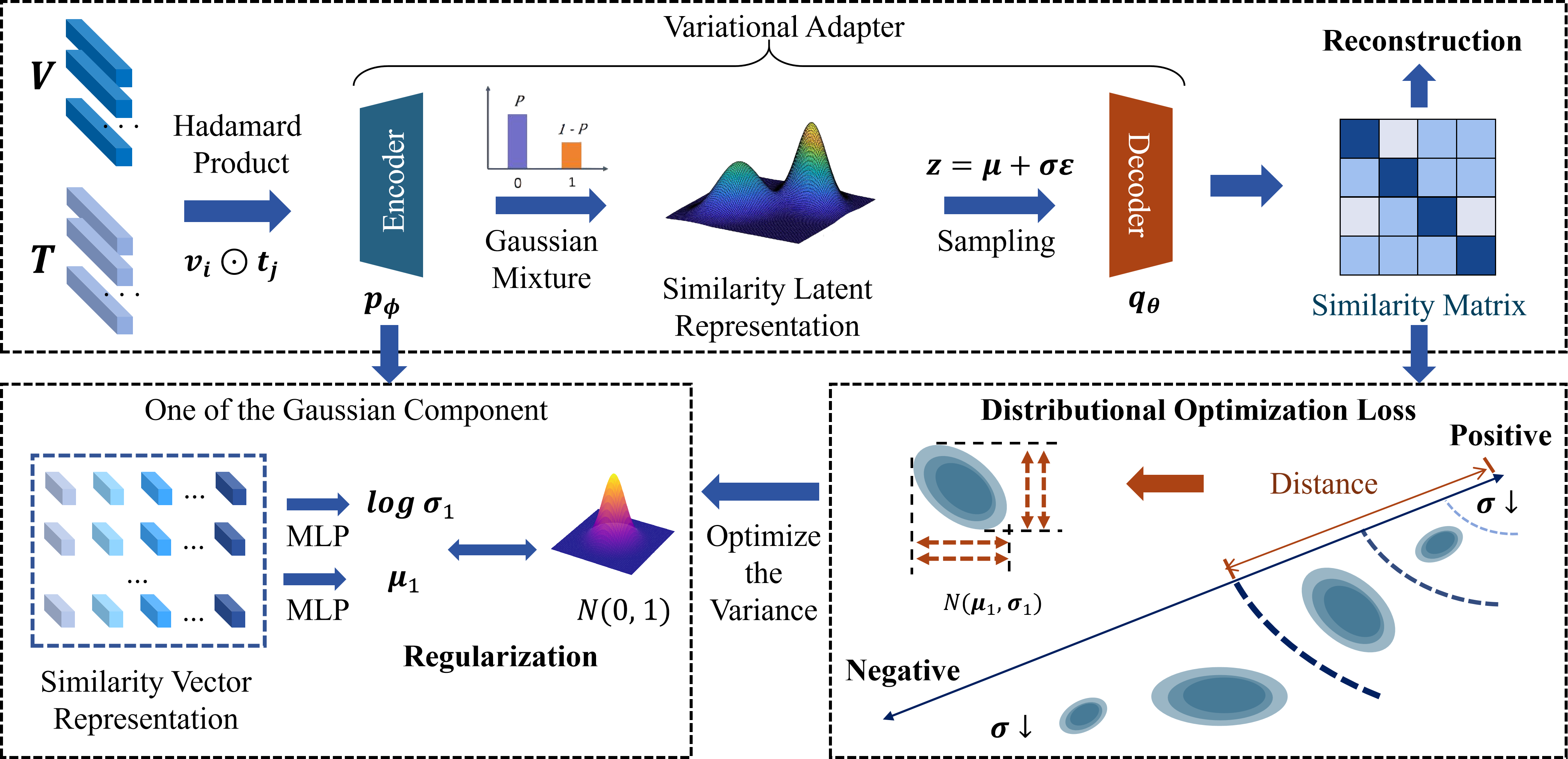}
\caption{Overview of our proposed model: Image and text features first interact through the Hadamard product to generate similarity vector representations, which are then input into a variational adapter composed of an encoder and a decoder. The encoder predicts the mean ($\mu$) and log-variance ($\text{log}\sigma^{2}$) for each similarity vector, mapping the input to a Gaussian mixture latent distribution, where each Gaussian component is regularized to a standard normal distribution. Using the reparameterization trick, latent variables are sampled and subsequently reconstructed into a similarity matrix by the decoder. The model is optimized by minimizing the reconstruction loss ($\mathcal{L}_{recon}$) between the predicted similarities and the binary labels, while a distributional optimization loss ($\mathcal{L}_{\sigma}$) is introduced to adaptively calibrate the uncertainty of the latent representations.}
\label{fig: main}
\end{figure*}
As mentioned earlier, the binary annotation forcibly separates the continuous similarity space, which can lead to the emergence of FNs. We begin by evaluating how widely used pairwise loss functions (contrastive loss and sigmoid loss \cite{zhai2023sigmoid}) are affected by FNs. We compute the gradient of the pairwise loss $\mathcal{L}(\mathbf{S})$ with respect to the output $B \times B$ similarity matrix $\bm{S}$ at the $t$-th iteration:
\begin{equation}
\begin{aligned}
    \frac{\partial \mathcal{L}(\bm{S})}{\partial S_{i,j}} &= \sum_{i=1}^{B}[(\sum_{j \neq i,(i,j)\in c}^{B}\frac{\partial \mathcal{L}(\bm{S})}{\partial S_{i,j}}+\sum_{j = i}^{B}\frac{\partial \mathcal{L}(\bm{S})}{\partial S_{i,j}}) \\
    &+\sum_{j \neq i,(i,j)\notin c}^{B}\frac{\partial \mathcal{L}(\bm{S})}{\partial S_{i,j}}]
\end{aligned}
\label{eq:1}
\end{equation}
where $S_{i,j}\in c$ denotes that the image text pair $(X_{i},Y_{j})$ belongs to the set of FNs. In practice, FNs exhibit potential semantic relations to positive samples yet provide gradients in the opposite direction. 
Following the definition of the relative penalty on negative samples in \cite{wang2021understanding}, we further extend this notion to quantify the gradient difference between positive and FNs as $r_{i} = |\sum_{j = i}\limits^{B}\frac{\partial \mathcal{L}(\bm{S})}{\partial S_{i,j}}| - |\sum_{j \neq i,(i,j)\in c}\limits^{B}\frac{\partial \mathcal{L}(\bm{S})}{\partial S_{i,j}}|$, which reflects the tolerance of the loss function to FNs. A smaller $r_{i}$ indicates greater difficulty for the model in learning semantically consistent representations. Meanwhile, the gradient of positive samples is also suppressed. When $r_{i} < 0$ the positive samples can no longer constrain the optimization direction of FNs, and semantically similar samples are forcibly separated, thereby disrupting the underlying semantic structure.

When the pair-based loss adopts contrastive loss $\mathcal{L}_{contrast} = -\log\frac{exp(S_{i,i}/\tau)}{\sum_{j\neq i}^{m}exp(S_{i,j}/\tau)+exp(S_{i,i}/\tau)}$, where $\tau$ represents the temperature coefficient, the corresponding $r_{i}$ can be computed as follows:
\begin{equation}
    r_{i}=\frac{1}{\tau} \frac{\sum_{j \neq i, (i,j) \notin c} exp(S_{i,j}/\tau)}{\sum_{j\neq i}^{n}exp(S_{i,j}/\tau)+exp(S_{i,i}/\tau)}
\label{eq:2}
\end{equation}
From Equation~\ref{eq:2}, we draw the following conclusions: (1) Since $r_{i}$ is always greater than 0, $\mathcal{L}_{contrast}$ does not completely disrupt the semantic consistency of representations; (2) The magnitude of $r_{i}$ equals the sum of all softmax-normalized negative samples multiplied by the reciprocal of $\tau$. Here, $\tau$ controls the smoothness of the distribution, a smaller $\tau$ causes an rapid reduction in the contribution of negative samples \cite{wang2021understanding}, thereby increasing the disruption to semantic consistency; a larger $\tau$ increases $r_{i}$ but can hinder the learning of separable features \cite{wang2021understanding}. Thus, $\mathcal{L}_{contrast}$ must balance $r_{i}$ and $\tau$.

When the pair-based loss adopts sigmoid loss: $\mathcal{L}_{sigmoid} = -\log\frac{1}{1 + exp(z_{i,j}(-aS_{i,j}+b))}$, where $z_{i,j}=1$ for positive samples and $z_{i,j}=-1$ otherwise. The corresponding $r_{i}$ can be computed as follows:
\begin{equation}
    r_{i}=|a| [(1 - P_{i,i}) - \sum_{S_{i,j}\in c} P_{i,j}]
\label{eq:3}
\end{equation}
where $P_{i,j}=sigmoid(-aS_{i,j}+b)$. In this case, the $r_{i}$ can be less than 0, $\mathcal{L}_{sigmoid}$ carries the risk of losing semantic information. Thus, it is necessary to employ suitable scaling parameters $a$ and $b$, which aligns with the empirical observations in SigLIP \cite{zhai2023sigmoid}. In Appendix \ref{TLF}, we compare the sensitivity of the contrastive loss and sigmoid loss to temperature coefficients and scaling parameters, further elucidating the impact of binary annotations on retrieval performance.

The foregoing analysis indicates that within the binary sparse annotation framework—whether using contrastive loss or sigmoid loss—the model struggles to fully mitigate semantic information loss due to FNs without resorting to additional hyperparameter tuning.
To address this limitation, we design a lightweight VAE-based probabilistic adapter that fine-tunes VLMs by explicitly modeling cross-modal similarity in a latent space. Owing to the variational autoencoder formulation, the adapter naturally introduces an MSE-based reconstruction loss under a Gaussian likelihood assumption. This loss function enables uncertainty-aware adaptive weighting of different image-text pairs, thereby reducing the gradient difference $r_{i}$.

%% file: section/3_methods.tex
\section{Methodology}
The overall architecture of VACSR is illustrated in Figure~\ref{fig: main}. We will detail the cross-modal similarity representation structure in Section \ref{se}, and elaborate on how to optimize the latent variable distribution for better modeling of FNs in Section \ref{opd}.
\subsection{Similarity Representation}
\label{se}
We propose a variational adapter to model the representation of cross-modal similarity, uncovering rich matching relationships that be overlooked by binary annotations.
As shown in Figure~\ref{fig: main}, a batch of image-text pairs are first encoded by a CLIP encoder to obtain their respective feature representations $\bm{V}=[\bm{v}_{1},\bm{v}_{2},...\bm{v}_{B}]\in\mathbb{R}^{B\times d}$ and $\bm{T}=[\bm{t}_{1},\bm{t}_{2},...\bm{t}_{B}]\in\mathbb{R}^{B\times d}$. Feature interaction is then achieved through the Hadamard product $s_{i,j} = \bm{v}_i \odot \bm{t}_j$, resulting in 
$\bm{S}\in\mathbb{R}^{B\times B\times d}$. 
where $\bm{s}_{i,j}\in \bm{S}$ can be regarded as a vector representation of similarity. Unlike the direct use of cosine similarity, the Hadamard product performs element-wise multiplication, which preserves the dimensionality of the interactive feature, thereby supporting subsequent encoding procedures. Subsequently, the latent representation $\mathbf{z}_{i,j}$ of $\bm{s}_{i,j}$ is constructed via the variational adapter. The design of this variational adapter references the variational autoencoder (VAE) \cite{kingma2013auto}, and we optimize the model by maximizing the Evidence Lower Bound (ELBO).
\begin{equation}
\begin{aligned}
\mathrm{ELBO}
&= \mathbb{E}_{p_{\phi}(\mathbf{z}_{i,j}\mid \bm{s}_{i,j})}
\left[
\log q_{\theta}(\hat{S}_{i,j}\mid \mathbf{z}_{i,j})
\right] \\
&\quad - \mathrm{KL}\left[
p_{\phi}(\mathbf{z}_{i,j}\mid \bm{s}_{i,j})
\,\middle\|\,
q(\mathbf{z}_{i,j})
\right]
\end{aligned}
\label{ELBO}
\end{equation}
where $\hat S_{i,j}$ denotes the true similarity containing fine-grained information, which we temporarily replace with the binary label $\hat{y}_{i,j}$. We will introduce how to correct $\hat{y}_{i,j}$ in Section \ref{opd}. Prior $q(\mathbf{z}_{i,j})$ is set to the standard normal distribution. $p_{\phi}$ and $q_{\theta}$ denote the parameterized similarity representation encoder and decoder composed of multilayer perceptrons (MLPs), respectively. 
To prevent the unimodal nature of a Gaussian distribution from restricting the model's capacity to learn complex semantic representations \cite{bai2022gaussian,ye2025towards}, we approximate $p_{\phi}(\mathbf{z}_{i,j}|\bm{s}_{i,j})$ as a two-component Gaussian mixture:
\begin{equation}
p_{\phi}(\mathbf{z}_{i,j}|\bm{s}_{i,j}) =
\sum_{k=1}^{2}\alpha_k \mathcal{N}\big(\mathbf{z}_{i,j}|\mu_k(\bm{s}_{i,j}), \operatorname{diag}(\sigma_k^2(\bm{s}_{i,j}))
\big),
\end{equation}
where $\mu_{k}, \sigma_{k}\in \mathbb{R}^{d}$ are derived from the $\mu$ and $\log\sigma^{2}$ heads corresponding to different Gaussian components in $p_{\phi}$, and $\alpha$ is learnable mixing coefficient with the constraint that $\alpha_k \ge 0$ and $\sum_{k=1}^{2}\alpha_k=1$.
The exact KL divergence between a Gaussian mixture posterior and the standard normal prior generally has no closed-form expression. Therefore, instead of computing the exact KL, we use a tractable variational upper bound derived from the convexity of KL divergence:
\begin{equation}
\begin{aligned}
& \text{KL}\left[\sum_{k=1}^{2}\alpha_k p_{\phi}^{k}(\mathbf{z}_{i,j}|\bm{s}_{i,j}) \middle\|\ q(\mathbf{z}_{i,j})\right]\\
&\le\sum_{k=1}^{2}\alpha_k
\text{KL}\left[p_{\phi}^{k}(\mathbf{z}_{i,j}|\bm{s}_{i,j}) \middle\|\ q(\mathbf{z}_{i,j})\right]
\end{aligned}
\end{equation}
Thus, the KL regularization term is implemented as the following conservative surrogate(A complete derivation is provided in Appendix~\ref{app:mixture_kl_bound}.):
\begin{equation}
\begin{aligned}
\mathcal{L}_{KL}
=
&\alpha_{1} \text{KL}\left[
\mathcal{N}(\mu_{1}(\bm{s}_{i,j}),\operatorname{diag}(\sigma_{1}^{2}(\bm{s}_{i,j})))
\,\middle\|\,
\mathcal{N}(\mathbf{0},\mathbf{I})
\right] \\
+
&\alpha_{2} \text{KL}\left[
\mathcal{N}(\mu_{2}(\bm{s}_{i,j}),\operatorname{diag}(\sigma_{2}^{2}(\bm{s}_{i,j})))
\,\middle\|\,
\mathcal{N}(\mathbf{0},\mathbf{I})
\right].
\end{aligned}
\end{equation}
The VAE approximates the generative model $q_{\theta}(\hat S_{i,j}|\mathbf{z}_{i,j})$ as a Gaussian distribution, and the reconstruction term $\mathcal{L}_{recon} = \mathbb{E}_{p_{\phi}(\mathbf{z}_{i,j}|\bm{s}_{i,j})}[\log q_{\theta}(\hat S_{i,j}|\mathbf{z}_{i,j})]$ is formulated as the negative log-likelihood under this Gaussian assumption, where:
\begin{equation}
\begin{aligned}
&\log q_{\theta}(\hat S_{i,j}|\mathbf{z}_{i,j})\\
&=\log(\frac{1}{\sqrt{2\pi \sigma^{2}(\mathbf{z}_{i,j})}}\exp(-\frac{(\hat{y}_{i,j} -\mu(\mathbf{z}_{i,j}))^{2}}{2\sigma^{2}(\mathbf{z}_{i,j})}
)) \\
& =\frac{1}{2\sigma^{2}(\mathbf{z}_{i,j})}||\hat y_{i,j} - \mu(\mathbf{z}_{i,j})||^{2} 
+ \log\sigma(\mathbf{z}_{i,j})+\frac{1}{2}\log 2\pi
\end{aligned}
\label{recons}
\end{equation}
Ignoring the constant term, this loss is equivalent to the mean squared error (MSE) loss. Here, $\mu(\mathbf{z}_{i,j})\in \mathbb{R}^{1}$ is obtained from the decoder $q_{\theta}$, normalized by a sigmoid function, and regarded as the final similarity score. The variance $\sigma^{2}(\mathbf{z}_{i,j})$ is typically fixed to 1. For the latent variable $\mathbf{z}_{i,j}$, we first use the Gumbel-Softmax trick to reparameterize the binary Gaussian component selection process, obtaining a differentiable approximate one-hot vector. We then apply the standard Gaussian reparameterization trick to sample from each Gaussian component. The final sampling point is obtained through a one-hot vector weighted sum of the sampling results using the one-hot vector.

\input{table/comparison_result_about_COCO}
\input{table/comparison_result_about_ECCV_CXC}

\subsection{Optimization of Uncertainty in Latent Variables}
\label{opd}
As mentioned in Section~\ref{sec:intro}, the binary annotation $\hat{y}$ lacks sufficient fine-grained semantic information. Consequently, directly substituting $\hat{S}_{i,j}$ with $\hat{y}$ can result in incorrect gradients for FNs. We address this issue by managing the uncertainty of the latent variable. Given that $\mathbf{z}_{i,j}$ is sampled from $p_{\phi}(\mathbf{z}_{i,j}|\bm{s}_{i,j})$, we expand the reconstruction term $\mathcal{L}_{recon}$ with reparameterization trick as follows:
\begin{equation}
\begin{aligned}
    &\frac{1}{2}\mathbb{E}_{p_{\phi}(\mathbf{z}_{i,j}|\bm{s}_{i,j})}[||\hat{y} - \mu(\mathbf{z}_{i,j})||^{2}] \\
    &=\frac{1}{2}||\hat{y} - \mu[\hat{\mu}(\bm{s}_{i,j})+\varepsilon \cdot \hat{\sigma}(\bm{s}_{i,j})]||^{2}
\end{aligned}
\end{equation}
here $\hat{\mu}(\bm{s}_{i,j}),\hat{\sigma}(\bm{s}_{i,j})$ are derived from the currently sampled Gaussian components of $p_{\phi}$, and sampling is performed only once. $\varepsilon$ is Gaussian random noise. Let us consider two extreme cases: when $\hat{\sigma}(\bm{s}_{i,j}) \to 0$, $\mathcal{L}_{recon}$ optimizes $\hat{\mu}(\bm{s}_{i,j})$ towards binary results. Conversely, when $\hat{\sigma}(\bm{s}_{i,j}) \to \infty$, $\hat{\mu}(\bm{s}_{i,j})$ becomes negligible, and $\mu[\hat{\mu}(\bm{s}_{i,j})+\varepsilon \cdot \hat{\sigma}(\bm{s}_{i,j})]$ turns into Gaussian random noise. In this case, $\mathcal{L}_{recon}$ no longer influences the gradient of $\hat{\mu}(\bm{s}_{i,j})$. 
Thus, we can control the optimization strength of the binary annotations by adjusting $\hat{\sigma}(\bm{s}_{i,j})$ and only decode the $\hat{\mu}(\bm{s}_{i,j})$ from the $\mu$ head in $p_{\phi}$ as the final similarity output.

\input{table/noise}
Based on this analysis, in the $t$ iteration, we should assign lower uncertainty to positive samples and informative negative samples (typically hard negative samples \cite{wang2021understanding}) to enhance prediction accuracy, while assigning higher uncertainty to FNs to mitigate the interference caused by erroneous annotations. Accordingly, we introduce an additional distributional optimization loss for the variance $\hat{\sigma}(\bm{s}_{i,j})$:
\begin{equation}
    \mathcal{L}_{\sigma} = \frac{1}{2\hat{\sigma}(\bm{s}_{i,j})}||\hat y - \mu(\mathbf{z}_{i,j})||^{2} + \log\hat{\sigma}(\bm{s}_{i,j})
\end{equation}
We truncate the gradient of the term $||\hat y - \mu(\mathbf{z}_{i,j})||^{2}$ and use it solely as a weighting coefficient to optimize the variance term. By differentiating $\mathcal{L}_{\sigma}$ and setting the derivative to zero, we obtain:
\begin{equation}
\begin{aligned}
\frac{\partial \mathcal{L}_{\sigma}}{\partial \hat \sigma(\bm{s}_{i,j})} &= - \frac{1}{\hat \sigma^{3}(\bm{s}_{i,j})}||\hat y - \mu(\mathbf{z}_{i,j})||^{2} + \frac{1}{\hat \sigma(\bm{s}_{i,j})}=0 \\
&\hat \sigma^{2}(\bm{s}_{i,j}) = ||\hat y -\mu(\mathbf{z}_{i,j})||^{2}
\end{aligned}
\end{equation}
This indicates that the optimized variance equals the squared distance between the model's output similarity and the true label. Samples closer to the label exhibit smaller variance, while those farther away have larger variance, thus achieving our goal of assigning different levels of uncertainty to different samples. Furthermore, since the loss only constrains the variance within the $[0,1]$ interval, we additionally apply a sigmoid function to normalize the output $\log\sigma^{2}$ from $p_{\phi}$. This approach extends the domain of variance to $[0,+\infty]$ while preserving its range. Then, we compute the loss for positive samples and hard negative samples separately: 
\begin{equation}
\begin{aligned}
    &\mathcal{L}_{\sigma}^{P} = \frac{1}{2\hat \sigma^{2}(\mathbf{s}_{i,i})}||1 - \mu(\mathbf{z}_{i,i})||^{2} + \log\hat \sigma(\mathbf{s}_{i,i}) \\
    &\mathcal{L}_{\sigma}^{N} = \max_{j,j \neq i}[\frac{1}{2\hat \sigma^{2}(\mathbf{s}_{i,j})}||1 - \mu(\mathbf{z}_{i,j})||^{2} + \log\hat \sigma(\mathbf{s}_{i,j})]
\end{aligned}
\end{equation}
Here, we still set $\hat y=1$ for hard negative samples. In fact, cross-modal alignment relies on hard negative samples to better distinguish between positive and negative pairs \cite{faghri2017vse++}. If $\hat y=0$ were used, hard negative samples with similarity close to 1 would be assigned high uncertainty, which is detrimental to model's optimization. A sensitivity analysis on the number of selected hard negatives is provided in Appendix~\ref{sec:hard_negative_sensitivity}.

\textbf{Objective function}: The final objective function is defined by optimizing a weighted sum of multiple losses, with hyperparameters $\alpha$, $\beta$ and $\gamma$ serving as weighting coefficients. In all experiments, the hyperparameters are uniformly set to $\alpha=0.0005, \beta = \gamma=1$:
\begin{equation}
    \mathcal{L} = \alpha \mathcal{L}_{KL} + 
    \beta [\mathcal{L}_{recon} + \gamma(\mathcal{L}_{\sigma}^{P} + \mathcal{L}_{\sigma}^{N})] 
\label{eq:final}
\end{equation}
Meanwhile, we adopt a straight-through estimator (STE) \cite{jacob2018quantization} to address the gradient anomaly arising from the application of the MSE loss in classification tasks. Details can be found in Appendix \ref{STE}.

%% file: table/comparison_result_about_COCO.tex
\begin{table*}[t]
\centering
\caption{
The performance comparison between our model and other approaches on the COCO dataset. We present the Recall@K and RSUM metric results, including both the 1K test setting (averaged over 5-fold test datasets) and the 5K test setting. The best results are highlighted in bold.
}
\footnotesize 
\renewcommand{\arraystretch}{0.9} 
\setlength{\tabcolsep}{4pt} 
\begin{tabular}{@{} l *{6}{c}  *{6}{c}  @{}} 
\toprule
\multirow{2}{*}{Method} 
& \multicolumn{6}{c}{1K Test Images} 
& \multicolumn{6}{c}{5K Test Images} \\ 
\cmidrule(lr){2-7} \cmidrule(lr){8-13}
& \multicolumn{3}{c}{Image-to-Text} 
& \multicolumn{3}{c}{Text-to-Image}
& \multicolumn{3}{c}{Image-to-Text} 
& \multicolumn{3}{c}{Text-to-Image} \\
\cmidrule(lr){2-4} \cmidrule(lr){5-7} \cmidrule(lr){8-10} \cmidrule(lr){11-13}
& R@1 & R@5 & R@10 & R@1 & R@5 & R@10 
& R@1 & R@5 & R@10 & R@1 & R@5 & R@10 \\
\midrule
\multicolumn{13}{@{}l}{\textit{\textbf{CLIP ViT-B/32}}} \\ \midrule
P2RM~ 
    & 78.6 & 96.1 & 98.6 & 67.5 & 92.3 & 96.8 
    & 56.6 & 83.5 & 90.9 & 46.3 & 74.9 & 84.4 \\
DAA~ 
    & 79.8 & 96.5 & 98.9 & 67.4 & 91.9 & 96.5  
    & 59.8 & 85.0 & 92.0 & 46.1 & 74.7 & 84.2  \\
PCME~ 
    & 80.1 & 96.6 & 98.7 & 67.6 & 92.1 & 96.9 
    & 59.9 & 85.8 & 92.3 & 46.1 & 75.0 & 84.6 \\ 
PCME++~ 
    & 81.6 & 97.0 & 99.0 & 69.2 & 92.8 & 97.1 
    & 62.1 & 87.0 & 93.0 & 48.1 & 76.5 & 85.4 \\
\rowcolor{blue!8}
 \textbf{VACSR}
   &  \textbf{84.2} &  \textbf{97.2} &  \textbf{99.0} &  \textbf{70.3} &  \textbf{93.3} &  \textbf{97.4} 
   &  \textbf{66.5} &  \textbf{88.3} &  \textbf{93.9} &  \textbf{49.8} &  \textbf{77.7} &  \textbf{86.3}  \\

\midrule %
\multicolumn{13}{@{}l}{\textit{\textbf{CLIP ViT-B/16}}} \\ \midrule
P2RM~ 
    & 78.3 & 96.2 & 98.7 & 69.2 & 93.0 & 97.2 
    & 56.8 & 84.3 & 91.5 & 48.1 & 76.6 & 85.7 \\
DAA~ 
    & 46.1 & 76.8 & 87.6 & 41.3 & 73.4 & 85.1  
    & 24.3 & 49.9 & 62.7 & 22.4 & 47.1 & 59.1  \\
PCME~ 
    & 83.6 & 97.7 & 99.3 & 72.0 & 93.9 & 97.7 
    & 65.3 & 89.2 & 94.5 & 51.2 & 79.1 & 87.5 \\ 
PCME++~ 
    & 85.3 & 97.9 & 99.3 & 73.4 & 94.4 & 97.8 
    & 68.7 & 90.1 & 95.0 & 53.4 & 80.3 & 88.3 \\
\rowcolor{blue!8}
 \textbf{VACSR}
   &  \textbf{87.4} &  \textbf{98.2} &  \textbf{99.4} &  \textbf{74.3} &  \textbf{94.6} &  \textbf{97.9} 
   &  \textbf{72.2} &  \textbf{91.1} &  \textbf{95.4} &  \textbf{54.5} &  \textbf{81.1} &  \textbf{88.7} \\
\bottomrule %
\end{tabular}

\label{tab:coco_comparison} 
\end{table*}

%% file: table/comparison_result_about_ECCV_CXC.tex
\begin{table*}[t]
\centering
\caption{
The performance comparison between our model and other approaches on the EC and CxC datasets. We present the Recall@K, R-P and mAP@R metric results, the best results are highlighted in bold.
}
\footnotesize 
\renewcommand{\arraystretch}{0.9} 
\setlength{\tabcolsep}{4pt}
\begin{tabular}{@{} l *{6}{c} *{6}{c} @{}} 
\toprule
\multirow{2}{*}{Method} 
& \multicolumn{6}{c}{ECCV Caption} 
& \multicolumn{6}{c}{CxC} \\ 
\cmidrule(lr){2-7} \cmidrule(lr){8-13} 
& \multicolumn{3}{c}{Image-to-Text} 
& \multicolumn{3}{c}{Text-to-Image}
& \multicolumn{3}{c}{Image-to-Text} 
& \multicolumn{3}{c}{Text-to-Image} \\
\cmidrule(lr){2-4} \cmidrule(lr){5-7} \cmidrule(lr){8-10} \cmidrule(lr){11-13}
& R@1 & R-P & mAP@R & R@1 & R-P & mAP@R
& R@1 & R@5 & R@10 & R@1 & R@5 & R@10 \\
\midrule
\multicolumn{13}{@{}l}{\textit{\textbf{CLIP ViT-B/32}}} \\ \midrule
P2RM~ 
    & 72.2 & 41.7 & 30.2 & 89.5 & 55.5 & 47.6
    & 58.1 & 85.5 & 92.2 & 48.4 & 77.2 & 86.2\\
DAA~ 
    & 75.9 & 42.3 & 31.2 & 88.1 & 55.7 & 47.3
    & 61.5 & 86.8 & 93.3 & 48.2 & 76.9 & 86.1\\
PCME~ 
    & 74.9 & 42.3 & 31.2 & 88.0 & 55.5 & 47.1
    & 61.5 & 87.5 & 93.5 & 48.0 & 77.3 & 86.4\\ 
PCME++~ 
    & 76.6 & 43.4 & 32.3 & 89.5 & 55.9 & 47.8
    & 63.5 & 88.4 & 94.0 & 50.1 & 78.5 & 87.1\\
\rowcolor{blue!8}
 \textbf{VACSR}
   & \textbf{80.6} & \textbf{43.8} & \textbf{33.2} & \textbf{90.4} & \textbf{56.3} & \textbf{48.1}
   & \textbf{67.8} & \textbf{89.6} & \textbf{94.9} & \textbf{51.6} & \textbf{79.7} & \textbf{88.0}\\
\midrule
\multicolumn{13}{@{}l}{\textit{\textbf{CLIP ViT-B/16}}} \\ \midrule
P2RM~ 
    & 72.9 & 42.2 & 30.6 & 88.5 & 56.8 & 48.8
    & 58.5 & 86.0 & 92.7 & 50.0 & 78.7 & 87.3\\
DAA~ 
    & 40.3 & 22.4 & 12.4 & 60.0 & 38.9 & 29.0
    & 26.4 & 53.9 & 67.1 & 24.3 & 50.3 & 62.5\\
PCME~ 
    & 79.1 & 44.0 & 33.2 & 89.5 & 56.5 & 48.7
    & 66.8 & 90.5 & 95.4 & 53.1 & 80.9 & 88.9\\ 
PCME++~ 
    & 81.6 & 45.1 & 34.5 & 91.4 & 57.2 & 49.7
    & 69.9 & 91.3 & 95.7 & 55.2 & 82.0 & 89.7\\
\rowcolor{blue!8}
 \textbf{VACSR}
   & \textbf{84.9} & \textbf{45.5} & \textbf{35.4} & \textbf{92.2} & \textbf{57.5} & \textbf{49.7}
   & \textbf{73.3} & \textbf{92.0} & \textbf{96.2} & \textbf{56.3} & \textbf{82.8} & \textbf{90.1}\\
\bottomrule
\end{tabular}
\label{tab:eccv_cxc_comparison}
\end{table*}

%% file: table/noise.tex
\begin{table*}[t]
\centering
\caption{Noisy correspondence results using the ViT-B/32 backbone are shown. All reported metrics represent the average performance over both image-to-text and text-to-image retrieval directions.}
\label{tab:ablation_study}
\footnotesize
\renewcommand{\arraystretch}{0.9}
\setlength{\tabcolsep}{5pt} 
\begin{tabular}{@{\hspace{5pt}}ccccccccccc@{}}
\toprule
\multirow{2}{*}{Noise Ratio} & \multirow{2}{*}{Method} & \multicolumn{3}{c}{ECCV Caption} & \multicolumn{1}{c}{CxC} & \multicolumn{3}{c}{COCO} \\
\cmidrule(lr){3-5} \cmidrule(l){6-6} \cmidrule(l){7-9}
& & mAP@R & R-P & R@1 & R@1 & 1K R@1 & 5K R@1 & RSUM \\
\midrule

\multirow{6}{*}{20\%} & VSE${\infty}$ & 37.0 & 46.3 & 79.7 & 53.6 & 72.0 & 51.8 & 518.6 \\
& DAA & 6.7 & 12.5 & 18.5 & 7.0 & 15.3 & 6.0 & 212.8 \\
& PCME & 37.6 & 47.6 & 79.2 & 50.6 & 70.3 & 48.7 & 520.7 \\
& NCR & 35.9 & 46.0 & 78.0 & 50.6 & 70.1 & 48.8 & 518.6 \\
& BiCro & - & - & - & - & 71.3 & - & 523.2 \\
& PCME++ & 37.7 & 47.6 & 80.0 & 52.2 & 71.6 & 50.4 & 524.6 \\
& NPC & - & - & - & - & 73.1 & 53.8 & 529.8 \\
\rowcolor{blue!8}
& VACSR & \textbf{40.1} & \textbf{49.6} & \textbf{83.9} & \textbf{58.7} & \textbf{76.4} & \textbf{57.1} & \textbf{539.0} \\
\midrule
\multirow{6}{*}{50\%} & VSE${\infty}$ & 18.0 & 28.5 & 43.7 & 20.7 & 39.2 & 19.1 & 394.1 \\
& DAA & 0.3 & 0.8 & 1.0 & 0.3 & 0.8 & 0.2 & 20.9 \\
& PCME & 35.2 & 45.5 & 75.7 & 46.3 & 66.6 & 44.4 & 508.0 \\
& NCR & 34.0 & 44.3 & 75.1 & 47.3 & 66.8 & 45.5 & 508.5 \\
& PCME++ & 35.7 & 45.8 & 76.3 & 47.4 & 67.6 & 45.5 & 511.0 \\
& NPC & - & - & - & - & 71.3 & 51.9 & 523.4 \\
\rowcolor{blue!8}
& VACSR & \textbf{39.5} & \textbf{49.1} & \textbf{82.8} & \textbf{57.2} & \textbf{75.1} & \textbf{55.6} & \textbf{534.2} \\
\bottomrule
\end{tabular}
\label{tab:noise}
\end{table*}

%% file: section/4_experiment.tex
\section{Experiment}
\subsection{Experimental Setup}
We assess the generalization performance of the VACSR model on two downstream tasks: image-text retrieval and out-of-distribution generalization. Specifically, for the image-text retrieval task, we additionally introduce a \textbf{noisy correspondence} setting. For out-of-distribution scenarios, we consider two types: \textbf{base-to-novel generalization} and \textbf{domain generalization.} Detailed experimental settings are provided in Appendix \ref{TaD} and \ref{ID}.

\subsection{Main Results}
\textbf{Image-Text Retrieval}:
Tables \ref{tab:coco_comparison} and \ref{tab:eccv_cxc_comparison} present the performance comparison between VACSR and other methods on the image-text retrieval task. We select representative baselines including P2RM (MM) \cite{wang2022point}, DAA (NeurIPS) \cite{li2022differentiable}, PCME (CVPR) \cite{chun2021probabilistic}, and PCME++ (ICLR) \cite{chun2024improved}, all of which mitigate the FNs by modeling intra-modal ambiguity. Notably, our method employs a simple two-layer MLP structure instead of the Transformer-based uncertainty prediction used in PCME++, reducing the total parameter count to only 48.5\%of PCME++ (A detailed analysis of computational efficiency, including memory usage, throughput, and chunk-based inference strategy, is provided in \textbf{Appendix~\ref{app:efficiency}}). Experimental results demonstrate that VACSR achieves the best performance across all evaluation metrics under different backbone networks.
Specifically, compared to PCME++, VACSR exhibits superior performance on the mAP@R and R@1 metrics of the EC dataset, validating its capability to understand diverse semantic information. Furthermore, VACSR achieves significant improvements in R@1 across all evaluated datasets. Taking ViT-B/32 as an example, it achieves improvements of 3.2\%, 7.1\%, 5.2\%, and 6.8\% on the respective datasets in the image-to-text retrieval direction. This indicates that VACSR can assign appropriate uncertainty to different types of samples, thereby ensuring retrieval accuracy.
Moreover, when scaling the backbone network from ViT-B/32 to ViT-B/16, VACSR maintains performance improvements without any adjustments to the variational adapter. This demonstrates that our model can effectively model the semantic distribution of FNs, unaffected by the complexity of the backbone.

\textbf{Noisy Correspondence}:
As shown in Table \ref{tab:noise}, by injecting varying proportions of Noisy Correspondence into the training data, we systematically assess the model’s adaptability and stability under noisy conditions \cite{huang2021learning}. Additionally, we include three representative Noisy Correspondence learning methods—NCR \cite{huang2021learning}, BiCro \cite{yang2023bicro}, and NPC \cite{zhang2024negative}.
Experimental results demonstrate that the proposed VACSR method exhibits superior noise robustness across various noise levels. Under a 20\% noise ratio, VACSR outperforms the next-best method PCME++ by 6.3\%, 4.2\%, and 4.9\% in mAP@R, R-P, and R@1 of the ECCV Caption dataset, respectively; achieves a 12.5\% improvement in R@1 on the CxC dataset; and delivers 4.5\%, 6.1\%, and 1.7\% gains over NPC in 1K R@1, 5K R@1, and RSUM on the COCO dataset. At a high noise ratio of 50\%, VACSR maintains leading performance, while other methods drop sharply. It achieves improvements of 10.6\%, 7.2\%, 8.5\%, 20.7\%, 5.3\%, 7.1\%, and 2.1\% over the next-best methods across datasets.
In summary, VACSR effectively mitigates the impact of noisy annotations and demonstrates strong robustness.
\input{table/comparison_result_about_DOMAIN_ADAPTION}
\input{table/comparison_average_OPEN_CLASS}
\input{table/ablation_result}
\begin{figure*}[htbp]
\centering
    \includegraphics[width=1\textwidth]{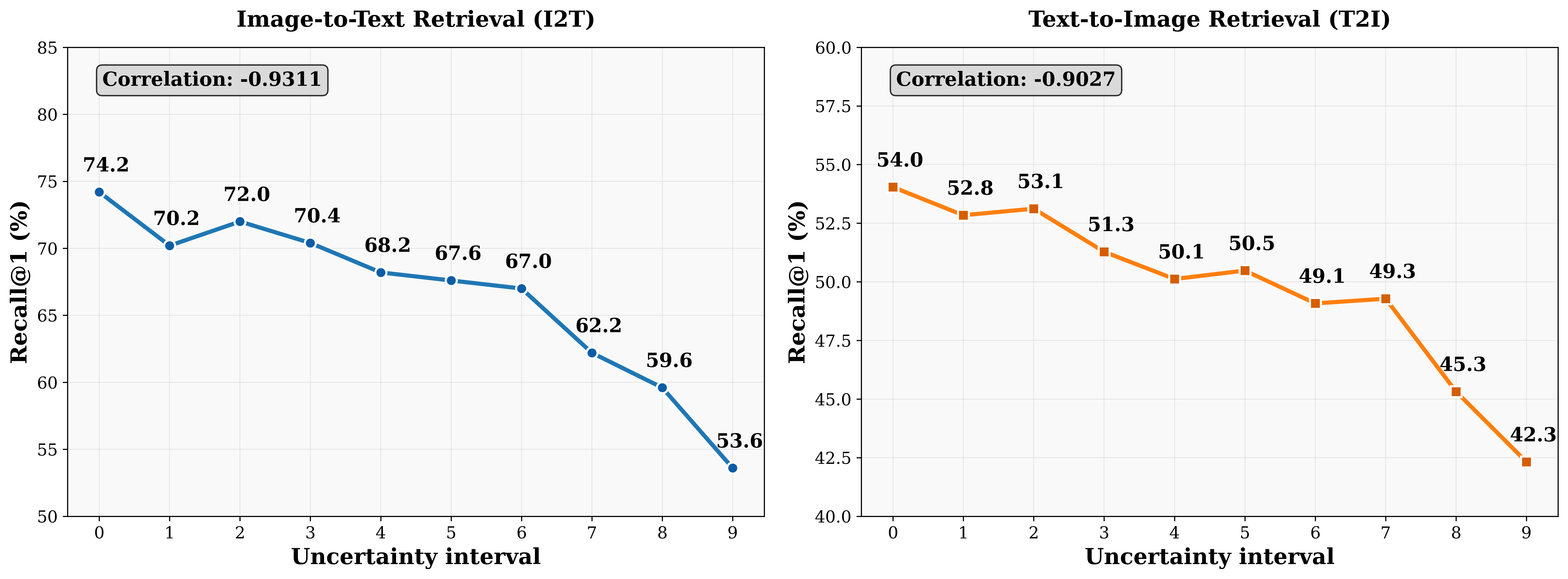}
    \caption{Visualization of relationship between uncertainty and R@1 metric.}
\label{fig: un_vs_r@1}
\end{figure*}

\textbf{Domain Generalization}:
CLIP extends its semantic understanding capability to classification tasks by computing the similarity between handcrafted text prompts (e.g., “a photo of a $<$class$>$”) and different visual categories
. However, classification datasets like ImageNet typically employ fixed annotations for visual categories, overlooking the potential fine-grained semantic relationships between different visual concepts and text. Therefore, we further conducted domain generalization experiments to evaluate the cross-task generalization capability of VACSR and the application potential of its similarity representation.
We compared our method with prompt-based learning approaches (e.g., COCOOP (CVPR) \cite{zhou2022conditional}, MaPLe (CVPR) \cite{khattak2023maple}, CoPrompt (ICLR) \cite{roy2023consistency}), adapter-based and fine-tuning methods (e.g., MMA (CVPR) \cite{yang2024mma}, CLIPOOD (ICML) \cite{shu2023clipood}), and methods that also model shared representation spaces (e.g., MMRL (CVPR) \cite{guo2025mmrl}). As shown in Table \ref{tab:domain_generalization_compariso}, VACSR achieved the best performance across all datasets, demonstrating its generalization capability and robustness to domain shifts. This result underscores that classification tasks also necessitate modeling a more continuous latent representation space to capture the inherent semantic of visual concepts effectively. 

\textbf{Base-to-Novel Generalization}:
The lack of semantic information in binary sparse annotations may also limit the model’s generalization ability to unseen categories. Therefore, we further conduct base-to-novel generalization experiments to evaluate whether similarity-based representations can learn more generalizable semantic features. 
As shown in Table \ref{tab:openclass_average}, VACSR achieves a harmonic mean (H) score of 80.37, outperforming most baseline methods (e.g., CoCoOp at 75.83 and MaPLe at 78.55), indicating strong overall performance. Notably, VACSR attains the best performance on base class recognition with a score of 85.74, demonstrating its advantage in retaining discriminative ability on seen categories.
By modeling cross-modal semantic distributions through variational inference, VACSR can partially overcome reliance on specific category labels, thus enabling generalization to semantically related unseen classes. Although there remains room for improvement compared to methods specifically designed for out-of-distribution generalization, these findings offer valuable insight into the relationship between similarity modeling and generalization ability. Please refer to Appendix \ref{BNG} for detailed results.

\subsection{Ablation Study}
\label{abst}
\textbf{Effectiveness of Each Component}
Table \ref{tab:ablation} demonstrates the effectiveness of $\mathcal{L}_{KL}$, $\mathcal{L}_{\sigma}$, the sigmoid function used in $\mathcal{L}_{\sigma}$, and the adoption of a Gaussian Mixture Model (GMM) prior. The baseline uses a Generalized Pooling Operator (GPO) and sigmoid loss to fine-tune CLIP directly. Experimental results show that the GMM prior has the most significant impact on overall performance, particularly in the EC dataset. This suggests that the unimodal nature of a single Gaussian prior is inadequate for fully modeling the latent space of similarity, further supporting the presence of complex semantic information in similarity representations. Compared to $\mathcal{L}_{KL}$, $\mathcal{L}_{\sigma}$ has a more pronounced effect on the EC datasets, indicating that $\mathcal{L}_{\sigma}$ effectively reduces the influence of binary annotations. Removing the sigmoid function leads to a decrease of 1.5\% in mAP@R and 1.4\% in R-P on the EC dataset. This phenomenon indicates that constraining the variance within a limited range weakens the model’s ability to express the degree of uncertainty.

\textbf{Metric Sensitivity under Uncertainty Modeling:}
Moreover, we observe that removing either $\mathcal{L}_{KL}$ or $\mathcal{L}_{\sigma}$ slightly increases R@1 on COCO and CxC. This result should be interpreted together with the different metric sensitivities. 
R@1 only considers the single highest-ranked ground-truth item and is highly sensitive to binary labels, whereas mAP@R and R-P evaluate ranking quality across all relevant items and are more robust to FNs. 
The two losses are designed to model similarity uncertainty and mitigate overfitting to binary labels. 
As a result, some FNs receive higher predicted similarity and may rank at the top, which slightly lowers R@1 on COCO and CxC under the official binary protocol.
On the cleaner and more densely annotated EC dataset, removing these two losses has little effect on R@1 but clearly degrades mAP@R and R-P, suggesting that they could improve ranking quality when the evaluation annotations better capture semantic relevance. This observation is consistent with prior findings that Recall@K can be sensitive to FNs \cite{chun2024improved}. Importantly, this does not imply that R@1 is unreliable. Under the standard evaluation protocol in Table \ref{tab:coco_comparison} and Table \ref{tab:eccv_cxc_comparison}, R@1 remains a valid measure of retrieving verified positives.

\subsection{Qualitative Analysis}
\textbf{Uncertainty and Accuracy}
Figure \ref{fig: un_vs_r@1} illustrates the correlation between uncertainty partitioning and retrieval accuracy. For the image-to-text retrieval task, we first compute the uncertainty $\hat{\sigma}(\bm{s}_{i,j})$ between each image and its most similar text as a confidence measure for that retrieval. Subsequently, all 5000 image retrieval results in COCO test datasets are divided into 10 equal-width intervals based on their confidence scores, and the R@1 accuracy within each interval is calculated. Each interval corresponds to a specific uncertainty range and its associated retrieval accuracy performance. A similar processing approach is applied to the text-to-image retrieval direction. As observed from the figure, both retrieval directions exhibit a significant negative correlation between uncertainty and R@1 (with correlation coefficients of -0.931 and -0.903, respectively). This result indicates that VACSR can assign reasonable uncertainty estimates to retrieval results: when uncertainty is high, the model exhibits lower confidence in its predictions, resulting in correspondingly lower retrieval accuracy; conversely, when uncertainty is low, model confidence is high, and retrieval accuracy improves significantly. Therefore, uncertainty can serve as an effective indicator of the reliability of retrieval results. This observation also underscores VACSR's strong interpretability. 
For more quantitative analyses, including ranking-level uncertainty evaluation, please refer to \textbf{Appendix~\ref{RUA}, ~\ref{SD} and ~\ref{VARR}}.

%% file: table/comparison_result_about_DOMAIN_ADAPTION.tex
\begin{table}[t]
\centering
\caption{Performance comparison of different methods on ImageNet and its variants. We employ Clip ViT-B/16 as the encoder backbone. The best results are highlighted in bold.}
\label{tab:imagenet_comparison}
\footnotesize 
\renewcommand{\arraystretch}{0.9} 
\setlength{\tabcolsep}{2pt}
\begin{tabular}{@{}lcccccc@{}}
\toprule
\multirow{2}{*}{METHOD} & \multicolumn{1}{c}{IN-DISTRIBUTION} & \multicolumn{5}{c}{OUT-OF-DISTRIBUTION} \\
\cmidrule(lr){2-2} \cmidrule(lr){3-7}
& IMAGENET & V2 & S & A & R & AVG. \\
\midrule
ZERO-SHOT & 66.7 & 60.8 & 46.1 & 47.8 & 74.0 & 57.2 \\
COCOOP & 71.0 & 64.1 & 48.8 & 50.6 & 76.2 & 59.9 \\
CLIPOOD & 71.6 & 64.9 & 49.3 & 50.4 & 77.2 & 60.4 \\
MaPLe & 70.7 & 64.1 & 49.2 & 50.9 & 77.0 &60.3 \\
CoPrompt & 70.8 & 64.3 & 49.4 & 50.5 & 77.5 & 60.4 \\
MMA & 71.0 & 64.3 & 49.1 & 51.1 & 77.3 & 60.5 \\
MMRL & 72.0 & 64.5 & 49.2 & 51.2 & 77.5 & 60.6 \\
\rowcolor{blue!8}
\textbf{VACSR}
   &  \textbf{74.3} &  \textbf{65.7} &  \textbf{49.7} &  \textbf{52.4} &  \textbf{78.4} &  \textbf{61.6} \\
\bottomrule
\end{tabular}
\label{tab:domain_generalization_compariso}
\end{table}

%% file: table/comparison_average_OPEN_CLASS.tex
\begin{table}[t]
\centering
\caption{Performance comparison of various methods on base-to-novel generalization across 11 datasets, using CLIP ViT-B/16 as the encoder backbone. Results are averaged over all 11 datasets.}
\footnotesize
\setlength{\tabcolsep}{12pt}
\renewcommand{\arraystretch}{0.9}
\begin{tabular}{lccc}
\toprule
 & BASE & NEW & H \\
\midrule
ZERO-SHOT & 69.34 & 74.22 & 71.70\\
\midrule
CoCoOp & 80.47 & 71.69 & 75.83\\
CLIPOOD & 83.9 & 74.5 & 78.9 \\
MaPLe & 82.28 & 75.14 & 78.55 \\
CoPrompt & 84.00 & \textbf{77.23} & 80.48 \\
MMA & 83.20 & 76.80 & 79.87 \\
MMRL & 85.68 & 77.16 & \textbf{81.20} \\
\rowcolor{blue!8}
VACSR & \textbf{85.74} & 76.08 & 80.37 \\
\bottomrule
\end{tabular}
\label{tab:openclass_average} 
\end{table}

%% file: table/ablation_result.tex
\begin{table*}[t]
\centering
\caption{Ablation study on the EC, CXC and COCO datasets. All reported metrics represent the average performance over both image-to-text and text-to-image retrieval directions.}
\footnotesize 
\renewcommand{\arraystretch}{0.9} 
\setlength{\tabcolsep}{5pt}
\begin{tabular}{@{\hspace{5pt}}lccccccc@{\hspace{5pt}}@{}}
\toprule
 & \multicolumn{3}{c}{ECCV Caption} & \multicolumn{1}{c}{CxC} & \multicolumn{3}{c}{COCO} \\
\cmidrule(lr){2-4} \cmidrule(lr){5-5} \cmidrule(lr){6-8} 
Configuration & mAP@R & R-P & R@1 & R@1 & 1K R@1 & 5K R@1 & RSUM \\
\midrule
w/ All Components & \textbf{40.7} & \textbf{50.1} & \textbf{85.5} & 59.7 & 77.2 & 58.1 & 541.4 \\
\midrule
w/o $\mathcal{L}_{KL}$ & 40.1 & 49.4 & 85.3 & 60.4 & 77.3 & 58.8 & \textbf{541.5} \\
w/o $sigmoid$ & 40.1 & 49.4 & 85.5 & 60.4 & 77.2 & 58.8 & 541.1 \\
w/o $\mathcal{L}_{\sigma}$ & 39.8 & 49.1 & 84.9 & \textbf{60.7} & \textbf{77.3} & \textbf{59.1} & 541.1 \\
w/o GMM & 39.6 & 49.0 & 84.1 & 59.8 & 77.0 & 58.1 & 540.7 \\
\rowcolor{blue!8}
w/o All Components (Baseline) & 39.3 & 48.7 & 83.1 & 57.3 & 75.5 & 55.6 & 537.0 \\
\bottomrule
\end{tabular}
\label{tab:ablation}
\end{table*}

%% file: section/5_conclusion.tex
\section{Conclusion}
This work focuses on addressing the false negative sample problem caused by binary annotations. Unlike previous methods that focus on the inherent ambiguity of data, VACSR directly learning the probabilistic representation of cross-modal similarity to capture fine-grained semantic information by variational inference. Experimental results indicates the broad potential of probabilistic similarity representation in various multimodal downstream tasks.

%% file: section/6_appendix.tex
\section{Related Work}
\label{DBVP}
Latent variable models\cite{kingma2013auto,schonfeld2019generalized,lin2020learning,yi2021cross} represent sparse one-to-one correspondences as probabilistic distributions in the latent space, providing explicit physical interpretations for different variables. This approach extends the range of semantic expression and captures semantic ambiguity.

In cross-modal tasks, the sparsity of binary annotations often fails to accurately capture semantic relationships, leading to a large number of FNs. To address this, researchers have attempted to learn latent variable distributions by optimizing the probabilistic likelihood of contrastive objectives, in order to model the semantic ambiguity present in the data. The underlying assumption is that, if latent variables follow a certain distribution, data can be mapped to a high-dimensional distributional space via an encoder or decoder, thereby enhancing the expressive power of embedding semantics—resulting in so-called “probabilistic embeddings.” For example, Chun et al.\cite{chun2021probabilistic} construct a richer embedding space to implicitly capture one-to-many correspondences, where uncertainty estimation can further assist model decision-making. Subsequent work\cite{chun2024improved} improves the measurement of probability distributions, refines the construction of probabilistic spaces for image-text retrieval, and applies these methods to uncertainty-based zero-shot classification prompt tuning or pretraining task \cite{ji2023map}. Upadhyay et al.\cite{upadhyay2023probvlm} estimate the embedding distribution of pretrained vision-language models via posterior estimation, avoiding the overhead of retraining large models. Wang et al.\cite{wang2022point} introduce a geometric representation from point to rectangle, retrieving more relevant points within rectangular regions to enhance semantic recall. Li et al.\cite{li2022differentiable} directly optimize diversity metrics through differentiable approximation functions, addressing the challenge of non-differentiable objective optimization. Collectively, these approaches expand the set of potential results, constructing a richer semantic retrieval space.

However, existing probabilistic embedding methods typically model images and texts as separate probability distributions. The underlying intuition is that the ambiguity in image-text matching relationships is equivalent to the inherent uncertainty of the image and text data themselves. This implies that even when the data semantic are deterministic, incorrect annotations can still lead to erroneous uncertainty predictions.
The core innovation of VACSR lies in directly representing the matching relationship between image and text data—that is, the cross-modal similarity—as a probability distribution, rather than modeling the distributions of the image or text modalities themselves. On this basis, we propose a distributional optimization loss that replaces the discrete similarity distribution implied by original binary annotations with a continuous, hypothesized probabilistic similarity distribution, thereby avoiding the interference of binary labels.

\section{Method Details}
\subsection{Straight-Through Estimation for MSE Loss}
\label{STE}
\begin{figure*}[htbp]
\centering
    \includegraphics[width=0.7\textwidth]{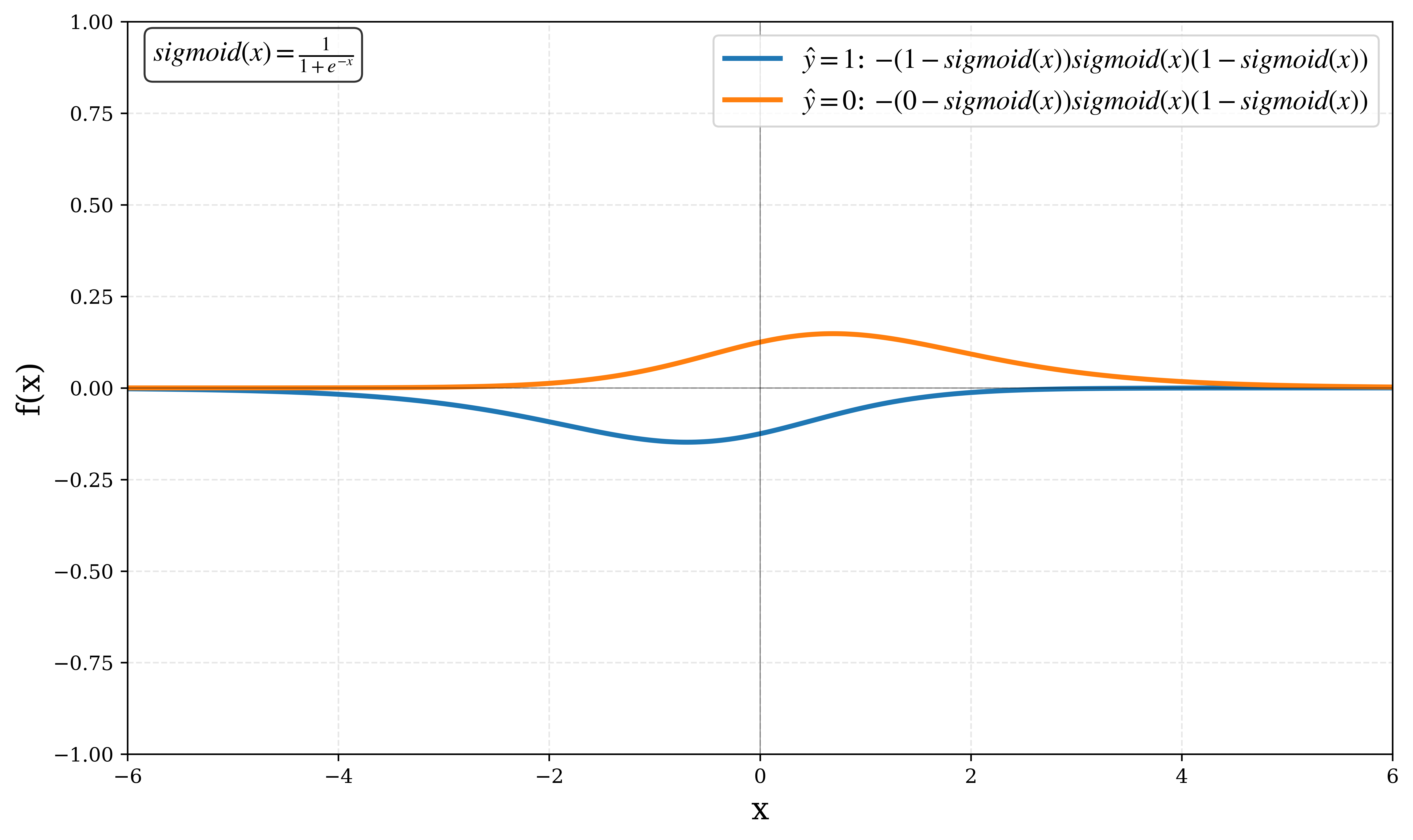}
    \caption{Visualization of MSE loss in classification task.}
\label{fig: mse_grad}
\end{figure*}
As described in Section \ref{se}, we adopt the mean squared error (MSE) loss as the optimization objective. This loss models the generative process as a Gaussian distribution, which simplifies the computation and ensures the continuity and structure of the latent space. However, since the annotations are binary, we need to apply sigmoid normalization to the decoder output $\mu(\mathbf{z}_{i,j})$. In this case, the derivative of Equation \ref{recons} with respect to $\mu(\mathbf{z}_{i,j})$ is:
\begin{equation}
    \frac{\partial ||\hat{y} - sigmoid(\mu(\mathbf{z}_{i,j}))||^{2}}{2\partial \mu(\mathbf{z}_{i,j})}=-||\hat{y} - sigmoid(\mu(\mathbf{z}_{i,j}))||sigmoid(\mu(\mathbf{z}_{i,j}))(1-sigmoid(\mu(\mathbf{z}_{i,j})))
\label{ste_par}
\end{equation}
The behavior of this gradient function is illustrated in Figure~\ref{fig: mse_grad}: regardless of the sample’s positive or negative nature, the gradient approaches zero when the similarity value is near 0 or 1, thereby hindering effective classification. To address this issue, we employ the Straight-Through Estimator (STE) to approximate the gradient, bypassing the differentiation of the sigmoid function by directly treating the output derivative as the input derivative. Thus, Equation~\ref{ste_par} becomes:
\begin{equation}
    \frac{\partial ||\hat{y} - sigmoid(\mu(\mathbf{z}_{i,j}))||^{2}}{2 \partial \mu(\mathbf{z}_{i,j})}=-||\hat{y} - sigmoid(\mu(\mathbf{z}_{i,j}))||
\end{equation}
This formulation is equivalent to the gradient of the sigmoid loss, thereby ensuring stable gradient propagation throughout the learning process.

\subsection{Derivation of the KL Upper Bound for the Gaussian Mixture Posterior}
\label{app:mixture_kl_bound}
In this section, we justify the KL regularization used for the two-component Gaussian mixture posterior. Let
\begin{equation}
p(\mathbf{z}) = \sum_{k=1}^{K}\alpha_k p_k(\mathbf{z}),
\quad \alpha_k \ge 0,\quad \sum_{k=1}^{K}\alpha_k=1
\end{equation}
where each component is a diagonal Gaussian
\begin{equation}
p_k(\mathbf{z})=\mathcal{N}(\mathbf{z}|\mu_k,\operatorname{diag}(\sigma_k^2))
\end{equation}
and the prior is the standard normal distribution
\begin{equation}
q(\mathbf{z})=\mathcal{N}(\mathbf{0},\mathbf{I})
\end{equation}

The exact KL divergence between the mixture posterior and the prior is
\begin{equation}
\text{KL}(p(\mathbf{z})\|q(\mathbf{z}))
=
\int
\left(
\sum_{k=1}^{K}\alpha_k p_k(\mathbf{z})
\right)
\log
\frac{
\sum_{k=1}^{K}\alpha_k p_k(\mathbf{z})
}{
q(\mathbf{z})
}
d\mathbf{z}
\end{equation}
This quantity generally has no closed-form expression because the entropy term
\begin{equation}
\begin{aligned}
-\int p(\mathbf{z}) \log p(\mathbf{z}) d\mathbf{z}  
= -\sum_{k=1}^{K}\alpha_k \int \mathcal{N} (\mathbf{z}|\mu_k,\operatorname{diag}(\sigma_k^2))  \log  \left(\sum_{j=1}^{K}\alpha_j \mathcal{N}  (\mathbf{z}|\mu_j,\operatorname{diag}(\sigma_j^2)) \right) d\mathbf{z}
\end{aligned}
\end{equation}
contains the logarithm of a mixture density.
However, the KL divergence is convex in its first argument. Therefore, for any set of distributions $\{p_k\}_{k=1}^{K}$ and non-negative mixture weights $\{\alpha_k\}_{k=1}^{K}$ satisfying $\sum_{k=1}^{K}\alpha_k=1$, we have
\begin{equation}
\mathrm{KL}\left(
\sum_{k=1}^{K}\alpha_k p_k(\mathbf{z})
\,\middle\|\,
q(\mathbf{z})
\right)
\le
\sum_{k=1}^{K}\alpha_k
\mathrm{KL}(p_k(\mathbf{z})\|q(\mathbf{z}))
\end{equation}
For completeness, we provide the proof below. Starting from the log-sum inequality, for non-negative functions $a_k(\mathbf{z})$ and $b_k(\mathbf{z})$,
\begin{equation}
\left(\sum_{k=1}^{K} a_k(\mathbf{z})\right)
\log
\frac{
\sum_{k=1}^{K} a_k(\mathbf{z})
}{
\sum_{k=1}^{K} b_k(\mathbf{z})
}
\le
\sum_{k=1}^{K}
a_k(\mathbf{z})
\log
\frac{
a_k(\mathbf{z})
}{
b_k(\mathbf{z})
}
\end{equation}
Setting
\begin{equation}
a_k(\mathbf{z})=\alpha_k p_k(\mathbf{z}),
\qquad
b_k(\mathbf{z})=\alpha_k q(\mathbf{z})
\end{equation}
we obtain
\begin{equation}
\left(\sum_{k=1}^{K}\alpha_k p_k(\mathbf{z})\right)
\log
\frac{
\sum_{k=1}^{K}\alpha_k p_k(\mathbf{z})
}{
\sum_{k=1}^{K}\alpha_k q(\mathbf{z})
}
\le
\sum_{k=1}^{K}
\alpha_k p_k(\mathbf{z})
\log
\frac{
\alpha_k p_k(\mathbf{z})
}{
\alpha_k q(\mathbf{z})
}
\end{equation}
Since $\sum_{k=1}^{K}\alpha_k q(\mathbf{z})=q(\mathbf{z})$ and the factors $\alpha_k$ cancel inside the logarithm on the right-hand side, this becomes
\begin{equation}
p(\mathbf{z})
\log
\frac{
p(\mathbf{z})
}{
q(\mathbf{z})
}
\le
\sum_{k=1}^{K}
\alpha_k p_k(\mathbf{z})
\log
\frac{
p_k(\mathbf{z})
}{
q(\mathbf{z})
}
\end{equation}
Integrating both sides over $\mathbf{z}$ gives
\begin{equation}
\mathrm{KL}(p(\mathbf{z})\|q(\mathbf{z}))
\le
\sum_{k=1}^{K}\alpha_k \mathrm{KL}(p_k(\mathbf{z})\|q(\mathbf{z}))
\end{equation}

Thus, replacing the exact mixture KL with the weighted sum of component-wise KL terms gives a tractable and conservative regularization term. Since the replacement upper-bounds the true KL, the resulting objective is a conservative lower-bound surrogate of the original ELBO and does not overestimate the variational objective.

\section{Experimental Details}
\subsection{Tasks and Datasets}
\label{TaD}
\textbf{Image-Text Retrieval}: Following \cite{chun2024improved}, we employ COCO Caption (COCO) \cite{chen2015microsoft} along with two extended benchmarks—ECCV Caption (EC) \cite{chun2022eccv} and CxC \cite{parekh2020crisscrossed} as the evaluation datasets. EC and CxC are built upon COCO with additional human annotations, significantly mitigating the FNs and providing a more reliable assessment of model generalization. Notably, EC corrects the largest number of FNs. We report Recall@K (R@K) for all benchmarks. For EC, we additionally report mAP@R and R-Precision (R-P) to provide a comprehensive evaluation of retrieval diversity and to more reliably reflect the model’s true generalization capability.

\textbf{Domain Generalization}: Following \cite{zhou2022conditional}, we trained VACSR on ImageNet \cite{deng2009imagenet} and evaluated on four variant datasets that introduce different domain shifts: ImageNet-V2 \cite{recht2019imagenet}, ImageNet-Sketch \cite{wang2019learning}, ImageNet-A \cite{hendrycks2021natural}, and ImageNet-R \cite{hendrycks2021many}. We employ a 16-shot setting and use the template "a photo of a $<$category$>$" for the word embeddings. This setup is used to assess the model’s generalization and robustness to out-of-distribution data.

\textbf{Base-to-Novel Generalization}: This evaluation follows widely adopted protocols \cite{zhou2022conditional,khattak2023maple}, aiming to assess the model's ability to recognize unseen categories after training on only a subset of classes. We conduct experiments on 11 image classification datasets spanning diverse recognition tasks, including general object recognition (ImageNet \cite{deng2009imagenet}, Caltech101 \cite{fei2004learning}), fine-grained recognition (OxfordPets \cite{parkhi2012cats}, StanfordCars \cite{krause20133d}, Flowers102 \cite{nilsback2008automated}, Food101 \cite{bossard2014food}, FGVCAircraft \cite{maji2013fine}), scene understanding (SUN397 \cite{xiao2010sun}), texture classification (DTD \cite{cimpoi2014describing}), satellite image recognition (EuroSAT \cite{helber2019eurosat}), and action classification (UCF101 \cite{soomro2012ucf101}). For each dataset, categories are evenly split into base and novel classes. The model is trained on the base classes using only 16 labeled samples per class in a few-shot setting, and subsequently evaluated on both base and novel test sets. This setup effectively measures the model’s adaptation on seen categories as well as its zero-shot generalization ability to unseen novel classes.

\subsection{Implementation Details}
\label{ID}
The core of our method lies in constructing probabilistic representations of cross-modal similarity rather than feature extraction itself; therefore, we adopt task-specific backbone networks for different tasks. For image-text retrieval, we follow the PCME++ framework \cite{chun2024improved}, utilizing a pre-trained CLIP as the backbone and employing the Generalized Pooling Operator (GPO) \cite{chen2021learning} for feature aggregation. To improve efficiency, we replace the original variance prediction Transformer module with a two-layer MLP-based variational adapter. Experiments are conducted with two visual encoders, ViT-B/32 and ViT-B/16, using the AdamP optimizer for 25 epochs, with an initial learning rate of 0.0005 decayed to 10\% at epoch 15.

For out-of-distribution generalization tasks, we follow the experimental setup of \cite{yang2024mma}, using the full CLIP model as the backbone and omitting structures such as GPO. In domain generalization, to prevent overfitting caused by the over-parameterization of CLIP and limited training samples, we fine-tune only the first two layers of the image encoder and the first three layers of the text encoder, training for 5 epochs with cosine annealing learning rate scheduling. For base-to-novel generalization, we conduct hyperparameter search on the number of frozen Transformer layers and training epochs, and adjust the batch size individually for datasets with extreme class distributions (128 for SUN397 and 5 for EuroSAT) to ensure training stability.

\section{Computational Efficiency Analysis}
\label{app:efficiency}
VACSR represents each image-text pair $(i,j)$ with a $d$-dimensional similarity vector, resulting in a tensor of size $B \times B \times d$ for a mini-batch of size $B$. While this design provides richer similarity modeling than scalar similarity scores, it may appear to introduce additional memory overhead. Nevertheless, the computational cost remains manageable, as the final objective is a pairwise MSE loss without global normalization. This allows each similarity vector $\bm{s}_{i,j}$ to be processed independently by the MLP encoder, enabling chunk-based computation in our implementation.
During inference, where the test set contains 5,000 images and 25,000 captions, we split the computation into $128 \times 128$ chunks along both dimensions. Each chunk requires a tensor of size $128 \times 128 \times 1024$, resulting in approximately 34 MB memory consumption, which is comparable to the training-time overhead. We further compare the inference efficiency of VACSR with the baseline PCME++ under the same hardware and batch size setting. All results are averaged over 100 runs. As shown in Table~\ref{tab:efficiency}, VACSR delivers consistent performance gains with substantially fewer parameters (48.5\%) and virtually no additional runtime cost.

\input{table/computational_efficiency}
\begin{figure*}[t]
\centering
\includegraphics[width=1\textwidth]{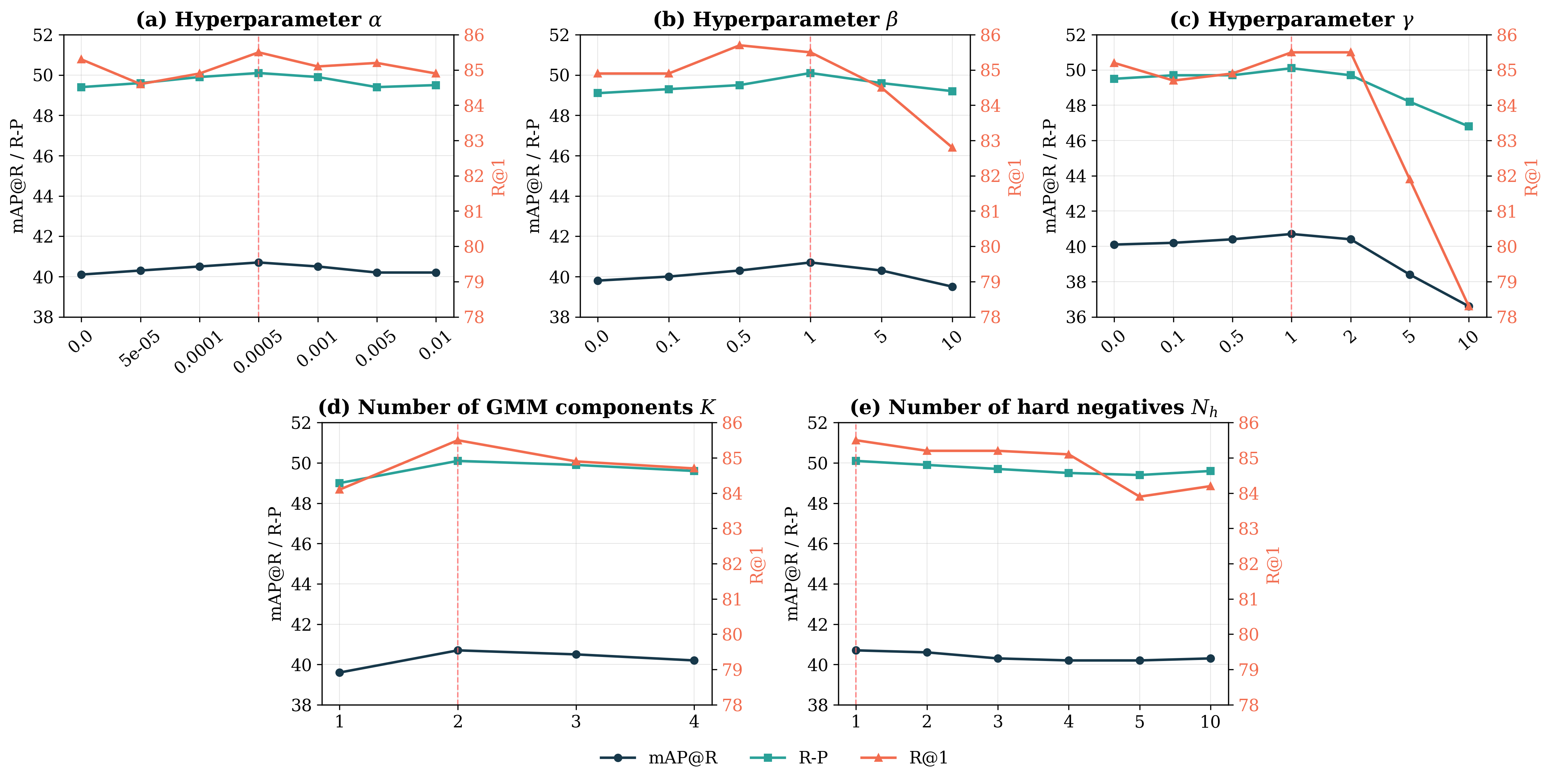}
    \caption{Sensitivity analysis on the EC dataset using the CLIP ViT-B/32 backbone. We evaluate the impact of the loss weighting coefficients \(\alpha\), \(\beta\), and \(\gamma\), the number of Gaussian components \(K\), and the number of selected hard negatives \(N_h\). All reported metrics are averaged over both image-to-text and text-to-image retrieval directions. The vertical dashed line in each subfigure indicates the hyperparameter value selected for our final model configuration.}
\label{fig: sensitive_analysis}
\end{figure*}
\input{table/contrastive_sigmoid_loss}
\section{Sensitivity Analysis}
\label{sana}
As shown in Figure~\ref{fig: sensitive_analysis}, we systematically analyze the impact of key parameters on model accuracy, including: (1) the loss function weighting coefficients \(\alpha\), \(\beta\), and \(\gamma\) in Eq.~\ref{eq:final}; (2) the number of components \(K\) in the Gaussian Mixture Model (GMM) described in Section~\ref{se}; and (3) the number of selected hard negatives \(N_h\). To isolate the effect of each variable, all experiments fix non-target parameters to their baseline values and vary only the parameter of interest. It is worth noting that the magnitude of \(\alpha\) is significantly smaller than that of the other parameters, such as \(\beta\) and \(\gamma\); therefore, a logarithmic scale is used for the search space of \(\alpha\) to ensure its effect is properly evaluated.

\subsection{The loss function weighting coefficients}
\textbf{Hyperparameter $\alpha$}: Analysis of the regularization weight $\alpha$ shows a clear performance peak around $\alpha$=0.0005. In addition, both excessively small (e.g., $\alpha$=0.00005) and large (e.g., $\alpha$=0.01) values of $\alpha$ lead to declines in all metrics. This indicates that appropriate regularization can improve retrieval performance, while overly strong regularization suppresses the model’s learning capacity. Notably, when regularization is completely removed (\(\alpha=0.0\)), the decrease in R@1 is relatively small (85.3), but mAP@R and R-Precision drop to 40.1 and 49.4, respectively, suggesting that regularization constrains the overall quality of retrieval results rather than merely improving Top-1 accuracy.

\textbf{Hyperparameter $\beta$}: Setting the weight $\beta$ of the distributional optimization loss to 1.0 yields the best trade-off among all metrics. When $\beta$ is too large (e.g., 10), the excessive emphasis on distributional optimization interferes with the primary learning objective, resulting in a significant drop in R@1 to 82.8. Conversely, when $\beta$ is too small (e.g., 0.1), the variance optimization is insufficient, and mAP@R decreases to 40.0. Additionally, at $\beta$=0.5, R@1 reaches a local optimum of 85.7, but both mAP@R and R-Precision decline, indicating that while the model emphasizes Top-1 accuracy, it may fail to maintain retrieval diversity.

\textbf{Hyperparameter $\gamma$}: The model achieves overall optimal performance when the weight $\gamma$ is set to 1.0, while an excessively large $\gamma$ (e.g., 10) leads to catastrophic performance degradation (mAP@R drops to just 36.6). This indicates that over-penalizing negative samples severely harms model performance. In fact, variance optimization for negative samples still uses positive labels, which preserves the model’s ability to learn from hard negatives but fails to capture precise uncertainty. This highlights the importance of appropriately weighting negative samples to enhance the model’s discriminative power.

In summary, analysis of the three parameters shows that the combination of $\alpha$=0.0005, $\beta$=1.0, and $\gamma$=1.0 achieves the best retrieval performance on the EC dataset (mAP@R 40.7, R-Precision 50.1, R@1 85.5). This configuration ensures sufficient regularization, balanced optimization of positive and negative samples, and avoids the negative effects of excessive penalization.

\subsection{The number of Gaussian components}
Sensitivity analysis of the component number $K$ demonstrates that this parameter critically affects the model’s ability to capture the intrinsic modal structure of the data. When $K=2$, the model achieves optimal performance on the EC dataset (mAP@R: 40.7, R-P: 50.1, R@1: 85.5), indicating that two Gaussian components effectively represent the similarity distribution patterns. When $K=1$, all metrics degrade significantly (e.g., mAP@R: 39.6) due to limited expressive capacity, suggesting that a single Gaussian is insufficient to model the data's complex characteristics. Conversely, when $K\geq3$, performance declines (with mAP@R at 40.2 when $K=4$), likely because the increased model complexity introduces redundant parameters, leading to overfitting on noisy training data and reduced generalization. As a result, $K=2$ is identified as the optimal choice, balancing sufficient expressiveness with the avoidance of over-parameterization risks.

\subsection{The number of selected hard negatives}
\label{sec:hard_negative_sensitivity}

Sensitivity analysis of the number of selected hard negatives \(N_h\) shows that the model achieves the best performance when only the single hardest negative is used, with mAP@R, R-P, and R@1 reaching 40.7, 50.1, and 85.5, respectively. As \(N_h\) increases, the performance decreases slightly and then remains stable, with mAP@R staying around 40.2--40.3 when \(N_h \geq 4\). This suggests that the method is not highly sensitive to the exact number of hard negatives, while introducing more negatives may include less FNs and slightly weaken the effectiveness of hard-negative optimization. Therefore, we set \(N_h=1\) as the default configuration.

\section{Tolerance of Loss Functions}
\label{TLF}
Table \ref{tab:loss} presents a comparative experiment between the sigmoid loss and the contrastive loss. For the sigmoid loss, $r_{i}$ does not require softmax normalization. When the similarity of positive samples is fixed, the tolerance of $\mathcal{L}_{sigmoid}$ depends solely on the FNs themselves. We observe that the sigmoid loss consistently outperforms the contrastive loss across all evaluation metrics, particularly in terms of R@K, indicating that its independence from softmax normalization during optimization contributes to higher retrieval accuracy. However, with fixed scaling parameters and temperature coefficients, the contrastive loss retains a baseline level of performance, while the sigmoid loss fails to converge, which further underscores the advantage of the contrastive loss in terms of robustness to binary annotations.

\section{Base-to-Novel Generalization}
\label{BNG}
\input{table/comparison_result_about_OPEN_CLASS}

This section provides a detailed analysis of the base-to-novel generalization results across 11 datasets, as shown in Table \ref{tab:openclass}, to further investigate the characteristics of VACSR on various recognition tasks.

\textbf{Generic Object Recognition Datasets (ImageNet, Caltech101):}
On ImageNet, VACSR achieves the highest base class accuracy (78.64) as well as the best harmonic mean (74.52), indicating strong fundamental performance for large-scale generic category recognition. On Caltech101, which contains a greater number of categories, VACSR ranks first in both novel class accuracy (95.09) and harmonic mean (96.89), demonstrating robust generalization to diverse object categories.

\textbf{Fine-grained Recognition Datasets (OxfordPets, StanfordCars, Flowers102, Food101, FGVCAircraft):}
For fine-grained tasks, different methods demonstrate varying strengths. VACSR achieves the highest base class accuracy on OxfordPets (96.28) and leads in novel class accuracy on Flowers102 (77.45). MMRL attains the best performance across base, novel, and harmonic mean metrics on StanfordCars, while CoPrompt slightly outperforms others in harmonic mean on Food101. On the highly challenging FGVCAircraft dataset, VACSR achieves the highest base class accuracy (47.96) and harmonic mean (41.55), indicating its advantage in learning representations for highly specialized categories.

\textbf{Scene, Texture, and Satellite Image Recognition (SUN397, DTD, EuroSAT):}
On the scene dataset SUN397, VACSR achieves the highest base class accuracy (83.23), while CoPrompt leads in novel class accuracy. For the texture dataset DTD, MMRL attains the best performance across multiple metrics. It is worth noting that on the EuroSAT satellite image dataset, the novel class accuracy of all methods is significantly lower than that of the base class. This is primarily due to the extremely limited textual category descriptions in this dataset (only five), which also severely constrains the effectiveness of approaches that rely on probabilistic representations (insufficient sampling) and results in a generalization bottleneck.

\textbf{Action Recognition Dataset (UCF101):}
On the UCF101 action recognition dataset, VACSR demonstrates a clear advantage, achieving the highest base class accuracy (89.04), novel class accuracy (80.31), and harmonic mean (84.45). This indicates its exceptional transfer capability in modeling dynamic semantic content.

Overall, VACSR shows robust base class recognition across most datasets and leads in comprehensive performance on datasets such as Caltech101, FGVCAircraft, and UCF101. Although its novel class generalization occasionally lags behind methods specifically designed for this purpose, its consistently strong and balanced performance validates the high generalizability of similarity representations learned via variational inference.
\begin{figure*}[t]
    \includegraphics[width=1.0\textwidth]{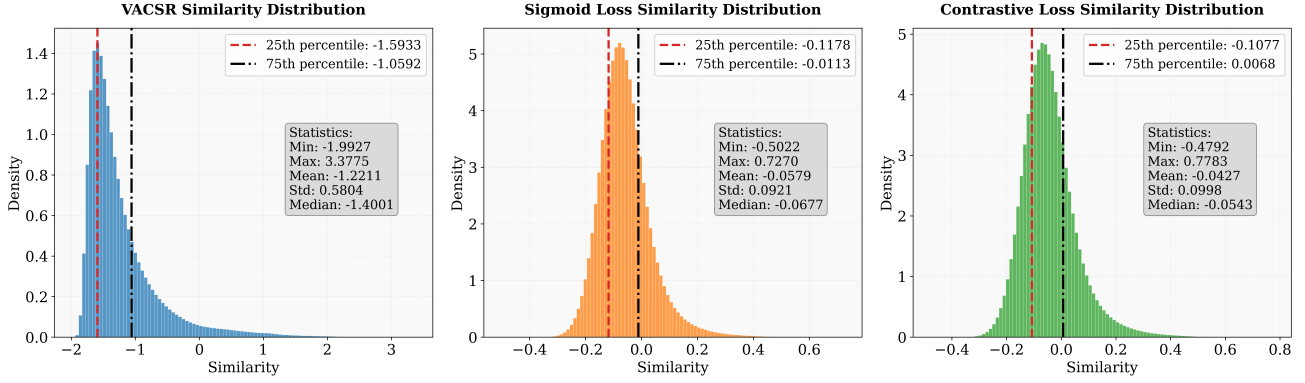}
    \caption{Visualization of similarity distribution.}
\label{fig: sim_distribution}
\end{figure*}

\input{table/rank_correlation}
\section{Ranking-level Uncertainty Analysis}
\label{RUA}
In the main paper, we analyze the relationship between uncertainty and retrieval accuracy using the uncertainty of the top-1 retrieved pair. Since image-text retrieval is inherently a list-level ranking problem, we further evaluate whether uncertainty reflects ranking quality beyond a single retrieved pair. Specifically, for each query, we compute the average uncertainty over its top-$K$ retrieved candidates:
\begin{equation}
    \bar{\sigma}_K = \frac{1}{K}\sum_{r=1}^{K} \hat{\sigma}(\bm{s}_{i,r}),
\end{equation}
where $\hat{\sigma}(\bm{s}_{i,r})$ denotes the predicted uncertainty of the $r$-th retrieved candidate for query $i$. We then report the Pearson correlation between $\bar{\sigma}_K$ and mAP@R on the EC dataset, where mAP@R is used as a ranking-sensitive metric.

These results in Table~\ref{tab:topk_uncertainty_mapr} yield several important insights.

\begin{itemize}
\item Firstly, strong negative correlations persist for small values of $K$ (top-5 in I2T and top-1 in T2I), confirming that a lower average uncertainty strongly predicts ranking quality precisely where it matters most.

\item Secondly, as $K$ increases, the correlation weakens and eventually becomes positive. This transition is expected: tail candidates naturally exhibit higher uncertainty, as they predominantly consist of FNs.
\end{itemize}

In our training data, the ratio of images to text is 1:5. Therefore, samples ranked before this ratio are considered positive; the lower their uncertainty, the higher the model’s confidence in the current retrieval result. In contrast, samples ranked after this ratio are generally treated as negative. In this case, the magnitude of uncertainty reflects the model’s confidence that “the sample may not actually be a negative sample” (i.e., whether it is a FN). The greater the uncertainty, the stronger the model’s belief that FNs exist (i.e. the better mAP@R). This trend is opposite to the direction of uncertainty change observed in positive samples. Meanwhile, because tail candidates inherently carry higher uncertainty, they dominate the average $\bar{\sigma}_K$ at large $K$.
These results further support the effectiveness of our uncertainty modeling. The predicted uncertainty serves as a reliable confidence indicator for highly ranked retrieval results, while also capturing ambiguity among lower-ranked candidates. 

\section{Similarity Distribution}
\label{SD}
Figure \ref{fig: sim_distribution} presents the similarity distributions obtained by VACSR and by fine-tuning CLIP directly using the sigmoid loss ($\mathcal{L}_{sigmoid}$) and the contrastive loss ($\mathcal{L}_{contrast}$). We take the COCO test set as an example, which contains $1.25\times10^{8}$ image-text pairs with a positive-to-negative sample ratio (including FNs) of $1:5000$. Figure \ref{fig: sim_distribution} shows the similarity distributions derived from $\mathcal{L}_{sigmoid}$ and $\mathcal{L}_{contrast}$ approximate a normal distribution (mean = -0.0579, -0.0427, std = 0.0921, 0.0998), indicating that the similarities of a large number of samples are concentrated in an intermediate range. As discussed in Section \ref{PRELIMINARY}, FNs are subjected to gradients in opposite directions and are thus pushed toward uncertain values rather than receiving confident judgments based on their true semantics. This result makes it difficult to distinguish FNs. In contrast, VACSR produces a markedly left-skewed distribution (mean = -1.2211, std = 0.5804). This distribution exhibits a lower central tendency (Median = -1.4001) and greater dispersion, demonstrating that VACSR can more fully span the similarities within the [0,1] interval, thereby effectively modeling the distribution of FNs. Furthermore, given the extremely small number of positive samples, the left-skewed distribution aligns more closely with our expectations regarding sample uncertainty. Specifically, the model can now assign lower similarity scores to the majority of negative samples with higher confidence, while reserving higher similarity values for the small number of positive samples. Consequently, VACSR alleviates the FNs caused by the binary annotations as a whole and achieves more fine-grained similarity modeling.

\section{Visualization Analysis of Retrieval Results}
\label{VARR}
\begin{figure*}[t]
\includegraphics[width=1.0\textwidth]{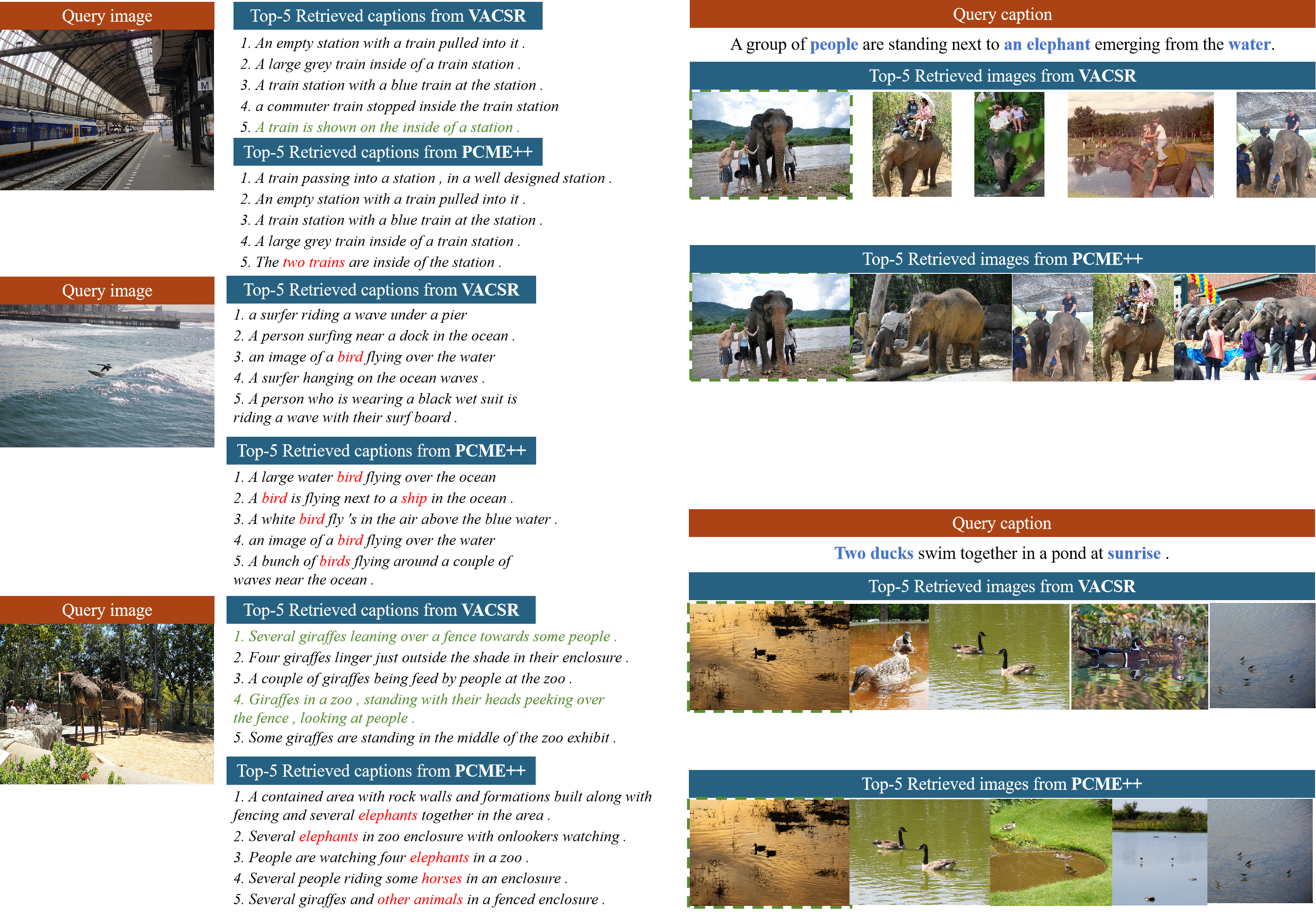}
    \caption{Visualization of retrieval results under a 50\% noise ratio. We show the top 5 results retrieved from each image or text query, with the positive samples labeled in the dataset boxed in green.}
\label{fig: qualitative_exp}
\end{figure*}
The visualization analysis of retrieval results under a 50\% noise ratio clearly demonstrates the robustness advantage of the VACSR method in real-world noisy environments. In the image-to-text retrieval task, VACSR is able to more accurately capture the core semantics of the query image. As shown in Figure \ref{fig: qualitative_exp}, for Query Image 1, all top-5 results retrievaled by VACSR closely align with the theme “A train is shown on the inside of a station,” whereas the results from PCME++ include semantic deviations such as “The two trains.” Similarly, in the cases of “surfer” and “giraffes,” VACSR’s results better maintain consistency with the main subject and scene of the query image, while PCME++ returns unrelated entities such as “birds,” “elephants,” and “people riding some horses.” This indicates that PCME++ is more susceptible to noise in the data and more likely to exhibit semantic drift.

In the text-to-image retrieval task, VACSR demonstrates a more precise grasp of the details in the query descriptions. For the text “A group of people are standing next to an elephant emerging from the water,” VACSR successfully retrieves multiple images containing the key element “water,” whereas most results from PCME++ overlook this detail. Similarly, for “Two ducks swim together in a pond at sunrise,” VACSR accurately matches the number of ducks, while PCME++ makes errors in this aspect. These examples indicate that by learning a distributional representation of similarity, VACSR is better able to distinguish key semantic features from noisy associations, thereby improving retrieval accuracy.

Notably, in both sets of experiments, many retrieved results were not labeled as “matched” but were highly semantically relevant to the queries. This phenomenon confirms the presence of numerous FNs in the original dataset. 

%% file: table/computational_efficiency.tex
\begin{table}[t]
\centering
\caption{Computational efficiency comparison between VACSR and the baseline PCME++.}
\footnotesize
\label{tab:efficiency}
\renewcommand{\arraystretch}{1} 
\setlength{\tabcolsep}{10pt}
\begin{tabular}{lccc}
\toprule
Metric & VACSR & PCME++ & Overhead \\
\midrule
Total Parameters & 158.56M & 327.07M & -51.5\% \\
Extra Module Parameters & 5.26M & 0M & +5.26M \\
Wall-clock Time per Step & 134.89 ms & 133.52 ms & +1.0\% \\
Throughput (img/s) & 949.0 & 958.7 & -1.0\% \\
Peak GPU Memory & 1.737 GB & 1.630 GB & +6.6\% \\
\bottomrule
\end{tabular}
\end{table}

%% file: table/contrastive_sigmoid_loss.tex
\begin{table}[h]
\centering
\caption{The impact of false-negative samples on the two types of losses. Here, ${L}_{sigmoid}$ and ${L}_{contrast}$ treat the scaling parameter and temperature coefficient as learnable parameters, initialized as $a=10,b=-10,\tau=1$. In contrast, ${L}_{sigmoid}^{\ast}$ and ${L}_{contrast}^{\ast}$ fix these parameters to $a=1,b=0,\tau=1$}
\label{tab:ablation_study}
\footnotesize 
\renewcommand{\arraystretch}{0.9} 
\setlength{\tabcolsep}{4pt}
\begin{tabular}{@{}lccccccccc@{}}
\toprule
 & \multicolumn{3}{c}{ECCV Caption} & \multicolumn{1}{c}{CxC} & \multicolumn{3}{c}{COCO} & \\
\cmidrule(lr){2-4} \cmidrule(lr){5-5} \cmidrule(lr){6-8}
Loss & mAP@R & R-P & R@1 & R@1 & 1K R@1 & 5K R@1 & RSUM \\
\midrule
VACSR & \textbf{40.7} & \textbf{50.1} & \textbf{85.5} & \textbf{59.7} & \textbf{77.2} & \textbf{58.1} & \textbf{541.4} \\
\midrule
${L}_{sigmoid}$ & 39.3 & 48.7 & 83.1 & 57.3 & 75.5 & 55.6 & 537.0 \\
${L}_{sigmoid}^{\ast}$ & 0.2 & 0.3 & 0.1 & 0.3 & 0.2 & 0.1 & 1.2 \\
${L}_{contrast}$ & 39.0 & 48.7 & 81.7 & 54.9 & 74.0 & 53.0 & 532.6 \\
${L}_{contrast}^{\ast}$ & 15.7 & 25.5 & 37.5 & 16.4 & 32.1 & 14.8 & 343.2 \\
\bottomrule
\end{tabular}
\label{tab:loss}
\end{table}

%% file: table/comparison_result_about_OPEN_CLASS.tex
\begin{table*}[t]
\centering
\caption{Performance comparison of different methods on base-to-novel generalization across 11 datasets. We employ Clip ViT-B/16 as the encoder backbone. The best results are highlighted in bold.}
\footnotesize
\setlength{\tabcolsep}{3.6pt}
\renewcommand{\arraystretch}{1.2}
\begin{tabular}{l|ccc|ccc|ccc|ccc}
\hline
\multirow{2}{*}{Method} & \multicolumn{3}{c|}{Average} & \multicolumn{3}{c|}{ImageNet} & \multicolumn{3}{c|}{Caltech101} & \multicolumn{3}{c}{OxfordPets} \\
& Base & Novel & HM & Base & Novel & HM & Base & Novel & HM & Base & Novel & HM \\
\hline
ZERO-SHOT & 69.34 & 74.22 & 71.70 & 72.43 & 68.14 & 70.22 & 96.84 & 94.00 & 95.40 & 91.17 & 97.26 & 94.12 \\
\midrule
CoCoOp & 80.47 & 71.69 & 75.83 & 75.98 & 70.43 & 73.10 & 97.96 & 93.81 & 95.84 & 95.20 & 97.69 & 96.43 \\
CLIPOOD & 83.9 & 74.5 & 78.9 & 77.5 & 70.3 & 73.7 & 98.7 & 94.6 & 96.6 & 95.7 & 96.4 & 96.0 \\
MaPLe & 82.28 & 75.14 & 78.55 & 76.66 & 70.54 & 73.47 & 97.74 & 94.36 & 96.02 & 95.43 & 97.76 & 96.58 \\
CoPrompt & 84.00 & \textbf{77.23} & 80.48 & 77.67 & 71.27 & 74.33 & 98.27 & 94.90 & 96.55 & 95.67 & \textbf{98.10} & \textbf{96.87} \\
MMA & 83.20 & 76.80 & 79.87 & 77.31 & 71.00 & 74.02 & 98.40 & 94.00 & 96.15 & 95.40 & 98.07 & 96.72 \\
MMRL & 85.68 & 77.16 & \textbf{81.20} & 77.90 & \textbf{71.30} & 74.45 & \textbf{98.97} & 94.50 & 96.68 & 95.90 & 97.60 & 96.74 \\
VACSR & \textbf{85.74} & 76.08 & 80.37 & \textbf{78.64} & 70.8 & \textbf{74.52} & 98.77 & \textbf{95.09} & \textbf{96.89} & \textbf{96.28} & 97.37 & 96.82 \\
\hline
\hline
\multirow{2}{*}{Method} & \multicolumn{3}{c|}{StanfordCars} & \multicolumn{3}{c|}{Flowers102} & \multicolumn{3}{c|}{Food101} & \multicolumn{3}{c}{FGVCAircraft} \\
& Base & Novel & HM & Base & Novel & HM & Base & Novel & HM & Base & Novel & HM \\
\hline
ZERO-SHOT & 63.37 & 74.89 & 68.65 & 72.08 & 77.80 & 74.83 & 90.10 & 91.22 & 90.66 & 27.19 & 36.29 & 31.09 \\
\midrule
CoCoOp & 70.49 & 73.59 & 72.01 & 94.87 & 71.75 & 81.71 & 90.70 & 91.29 & 90.99 & 33.41 & 23.71 & 27.74 \\
CLIPOOD & 78.6 & 73.5 & 75.9 & 93.5 & 74.5 & 82.9 & 90.7 & 91.7 & 91.2 & 43.3 & 37.2 & 40.0 \\
MaPLe & 72.94 & 74.00 & 73.47 & 95.92 & 72.46 & 82.56 & 90.71 & 92.05 & 91.38 & 37.44 & 35.61 & 36.50 \\
CoPrompt & 76.97 & 74.40 & 75.66 & 97.27 & 76.60 & 85.71 & \textbf{90.73} & \textbf{92.07} & \textbf{91.4} & 40.20 & \textbf{39.33} & 39.76 \\
MMA & 78.50 & 73.10 & 75.70 & 97.77 & 75.93 & 85.48 & 90.13 & 91.30 & 90.71 & 40.57 & 36.33 & 38.33 \\
MMRL & \textbf{81.30} & \textbf{75.07} & \textbf{78.06} & \textbf{98.97} & 77.27 & \textbf{86.78} & 90.57 & 91.50 & 91.03 & 46.30 & 37.03 & 41.15 \\
VACSR & 79.74 & 73.33 & 76.40 & 98.39 & \textbf{77.45} & 86.67 & 90.58 & 91.73 & 91.15 & \textbf{47.96} & 36.65 & \textbf{41.55} \\
\hline
\hline
\multirow{2}{*}{Method} & \multicolumn{3}{c|}{SUN397} & \multicolumn{3}{c|}{DTD} & \multicolumn{3}{c|}{EuroSAT} & \multicolumn{3}{c}{UCF101} \\
& Base & Novel & HM & Base & Novel & HM & Base & Novel & HM & Base & Novel & HM \\
\hline
ZERO-SHOT & 69.36 & 75.35 & 72.23 & 53.24 & 59.90 & 56.37 & 56.48 & 64.05 & 60.03 & 70.53 & 77.50 & 73.85 \\
\midrule
CoCoOp & 79.74 & 76.86 & 78.27 & 77.01 & 56.00 & 64.85 & 87.49 & 60.04 & 71.21 & 82.33 & 73.45 & 77.64 \\
CLIPOOD & 81.0 & 79.3 & 80.2 & 80.8 & 58.6 & 67.9 & \textbf{97.5} & 64.1 & 77.3 & 85.7 & 79.3 & 82.4 \\
MaPLe & 80.82 & 78.70 & 79.75 & 80.36 & 59.18 & 68.16 & 94.07 & 73.23 & 82.35 & 83.00 & 78.66 & 80.77 \\
CoPrompt & 82.63 & \textbf{80.03} & \textbf{81.31} & 83.13 & 64.73 & 72.79 & 94.60 & 78.57 & 85.84 & 86.90 & 79.57 & 83.07 \\
MMA & 82.27 & 78.57 & 80.38 & 83.20 & 65.63 & 73.38 & 85.46 & \textbf{82.34} & 83.87 & 86.23 & 80.03 & 82.20 \\
MM-RL & 83.20 & 79.30 & 81.20 & \textbf{85.67} & \textbf{65.00} & \textbf{73.82} & 95.60 & 80.17 & \textbf{87.21} & 88.10 & 80.07 & 83.89 \\
VACSR & \textbf{83.23} & 78.57 & 80.83 & 84.56 & 64.79 & 73.36 & 95.93 & 70.74 & 81.43 & \textbf{89.04} & \textbf{80.31} & \textbf{84.45} \\
\hline
\end{tabular}
\label{tab:openclass} 
\end{table*}

%% file: table/rank_correlation.tex
\begin{table*}[t]
\centering
\caption{Pearson correlation between the average top-$K$ uncertainty $\bar{\sigma}_K$ and mAP@R on the EC dataset.}
\footnotesize
\label{tab:topk_uncertainty_mapr}
\renewcommand{\arraystretch}{0.9} 
\setlength{\tabcolsep}{6pt}
\begin{tabular}{c|cccccccccc}
\toprule
$K$ & 1 & 2 & 3 & 4 & 5 & 6 & 7 & 8 & 9 & 10 \\
\midrule
I-T & -0.939 & -0.903 & -0.903 & -0.842 & -0.689 & -0.035 & 0.491 & 0.666 & 0.728 & 0.789 \\
T-I & -0.879 &  0.783 &  0.807 &  0.813 &  0.805 &  0.795 & 0.790 & 0.780 & 0.759 & 0.746 \\
\bottomrule
\end{tabular}
\end{table*}